\documentclass[11pt]{article}
\usepackage[preprint]{acl}
\usepackage{times}
\usepackage{latexsym}
\usepackage[T1]{fontenc}
\usepackage[utf8]{inputenc}
\usepackage{microtype}
\usepackage{graphicx}
\usepackage{booktabs}
\usepackage{array}
\usepackage{amsmath}
\graphicspath{{figures/}}

\title{The Neighbour's Label: Opposite Subgroup Effects of\\Same-Dialogue Exposure in Next-Act Evaluation}

\author{
  \textbf{Ali Amini}$^{1}$ \quad \textbf{Masoumeh Chapariniya}$^{2}$ \\
  \textbf{Rytis Maskeliūnas}$^{1}$ \quad \textbf{Robertas Damaševičius}$^{1}$ \\[3pt]
  $^{1}$Centre of Real Time Computer Systems, Kaunas University of Technology \\
  Kaunas, Lithuania \\
  $^{2}$Institute of Computational Linguistics, University of Zurich, Zurich, Switzerland \\[2pt]
  \texttt{ali.amini@ktu.edu} \quad \texttt{masoumeh.chapariniya@uzh.ch} \\
  \texttt{rytis.maskeliunas@ktu.lt} \quad \texttt{robertas.damasevicius@ktu.lt}
}

\begin{document}
\maketitle

\begin{abstract}
Window-based next-act evaluation often aims to measure generalisation across dialogues, yet overlapping windows and
similarity retrieval can expose labelled examples from the test dialogue. The nearest such example is the target's
labelled predecessor, whose act is correct when the target continues it and wrong when the target changes act; a model
that copies it can gain and lose in two subgroups that one aggregate averages. We induce this exposure under
controlled designs on SGD, STAR and AMI, whose official splits are clean. Randomly including the predecessor's row in
training raises a lexical classifier's continuation scores and lowers its transition scores in all three corpora, and
purging the adjacent rows removes most of the gain. With TF-IDF-retrieved demonstrations, two 7B
instruction-tuned models show the same opposite pattern in all eight model--corpus--sample arms (some single-cell
intervals include zero; one replication is partial), while natural-weight accuracy changes by only $-0.05$ to $+0.04$.
Exposure frequency depends on the retrieval pipeline. We evaluate two diagnostic comparisons without model retraining
and find that excluding the test dialogue does not detectably reduce accuracy on SGD. All exposure is induced; no
published system was re-evaluated.
\end{abstract}

\section{Introduction}

Models that predict the next dialogue act, intent, or strategy are trained and evaluated on examples that pair a window
of preceding turns with the act of the turn that follows. Consecutive windows of one dialogue share most of their text,
and the label of one window's target is the \emph{previous act} of the next window's target. When an evaluation aims to
measure generalisation to unseen dialogues (Setting~A, Section~\ref{sec:setting}), no gold label of a test dialogue may
reach the training set or the labelled demonstration pool; yet a row-level split places the window preceding a test
target in training with probability equal to the training share, and similarity retrieval over a pool of windows often
returns it. A prospective evaluation within a dialogue (Setting~B) may use earlier acts only if they are observed at run
time. Among the 78 accessible context-window dialogue studies we reviewed (2019--2026), 4 reported a window-level split
and 22 did not clearly state their split unit (Appendix~\ref{app:audit}).

\begin{figure*}[t]
\centering
\includegraphics[width=\textwidth]{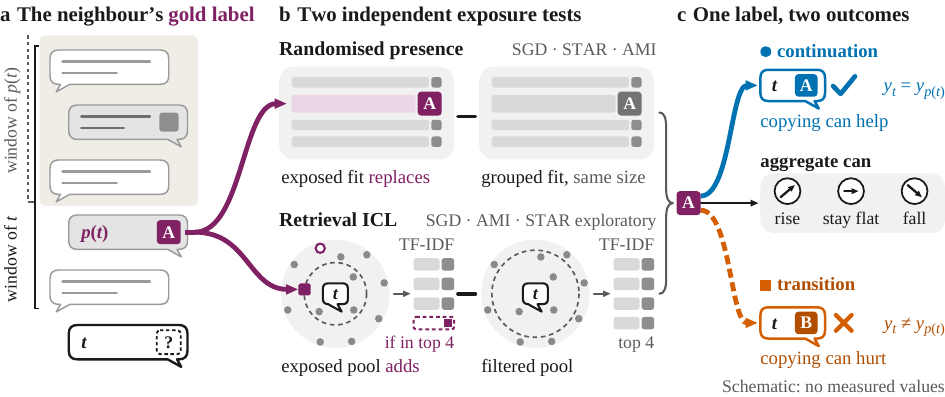}
\caption{Graphical overview (schematic, not measured values). (a) The labelled predecessor $p(t)$ can expose its gold label; earlier unlabelled dialogue text is not itself gold-label leakage. (b) Independent tests: randomly controlled, equal-size training-set replacement (SGD, STAR, AMI), and TF-IDF top-four retrieval from exposed versus filtered demonstration pools (SGD, AMI; exploratory STAR). (c) Copying $y_{p(t)}$ can help continuations, hurt transitions, and mask subgroup effects in the aggregate. All exposure is induced on officially clean splits; the dense retriever was used for prevalence only.}
\label{fig:overview}
\end{figure*}

\begin{figure}[t]
\centering
\includegraphics[width=\columnwidth]{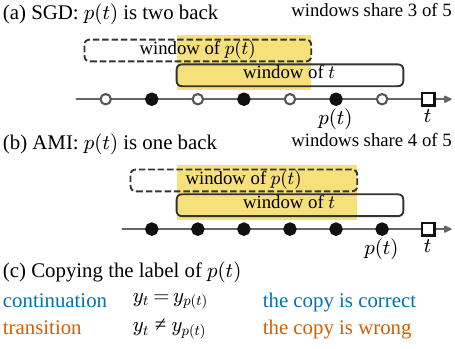}
\caption{(a, b) Labelled predecessor $p(t)$ of target $t$ (square) and the five-utterance windows of $t$ and $p(t)$;
shaded: utterances in both; filled: labelled turns, open: unlabelled user turns. STAR: two back for most targets.
(c) Copying the label of $p(t)$. Schematic; no corpus text.}
\label{fig:mechanism}
\end{figure}

The leaked item is the \emph{neighbour's} label, not the target's own, and its effect has a sign that depends on the
target (Figures~\ref{fig:overview} and~\ref{fig:mechanism}): when the target repeats the previous act (a \emph{continuation}) the neighbour's
label is the correct answer, and when it changes act (a \emph{transition}) it is wrong. A model that copies the label
gains on continuations and loses on transitions, and an aggregate of the same score on the same targets mixes the two
opposite terms in the corpus's proportions (Equation~\ref{eq:agg}), so it can rise, fall or stay flat while the
model's behaviour has changed. An aggregate below a published one therefore does not by itself indicate a cleaner
protocol, nor one above it leakage.

Each ingredient of this mechanism is documented. Language models copy the label of a demonstration that is
near-identical to the test input, even when it is wrong \citep{lyu2023zicl}, and their accuracy rises with the share of
demonstrations that carry the query's label \citep{chen2026rethinking}. The closest analogue to our
task is next-activity prediction in process mining, where most test prefixes duplicate training prefixes under standard
splits, a source of overestimated performance \citep{pfeiffer2025learning}; there the duplicated item is the test
input itself, here a neighbouring example whose label is right for one subgroup and wrong for the other. A retrieval-based dialogue-act annotation pipeline already excludes the
target session from retrieval as a precaution \citep{lee2026domain}, without reporting what the exclusion changes. What
remains unmeasured is the size and sign structure of this exposure in next-act evaluation, how often retrieval produces
it, whether it can be detected without retraining, and what removing it costs.

We measure these quantities by inducing the exposure on three public corpora whose official splits are dialogue- or
meeting-disjoint, SGD, STAR and AMI. A randomised presence design
isolates the adjacent labelled row for a lexical classifier, with placebo, purge and label-permutation controls and a
fine-tuned encoder as a second learner. Two 7B instruction-tuned models receive four demonstrations retrieved by TF-IDF
similarity from a pool that contains or excludes the test dialogue, on a first and a fresh query sample; a fixed pretrained dense
encoder serves as a second retrieval pipeline for exposure frequency only. Confirmatory decision rules were fixed before
their outcomes were observed; the exploratory follow-ups we report were designed after earlier results. Our
contribution is thus a measurement rather than a new mechanism, and its implication for claims about unseen dialogues
(Section~\ref{sec:discussion}) is to split by dialogue before windowing, exclude the test dialogue from demonstration pools, and report
continuation and transition cells next to the aggregate.

\section{Setting and Related Work}
\label{sec:setting}

\paragraph{Notation.} A dialogue $d$ is a sequence of utterances (in AMI, dialogue-act segments). A labelled target
$t$ is a system turn (SGD), a wizard turn with a selected action (STAR) or any segment (AMI); its input $x_t$ is the
five preceding utterances, and its gold label $y_t$ is a set of SGD acts, one of 161 STAR actions or one of 15 AMI
dialogue-act types. Let $p(t)$ be the previous labelled target of the same dialogue, which need not be the previous
utterance: it lies two utterances back in SGD, two in most STAR targets (four to sixteen in the rest) and one segment
back in AMI, so full-length windows of $t$ and $p(t)$ share three of five utterances in SGD and four in AMI. The target
is a continuation if $C_t=\mathbf{1}[y_t=y_{p(t)}]=1$, with exact equality of acts or act sets, and a transition
otherwise; dialogue-initial targets have no $p(t)$ and belong to neither cell. For an exposure effect measured with one
score on one set of targets, let $\Delta_C$ and $\Delta_T$ be its means over continuations and transitions and $\pi_C$
the share of continuations among non-initial targets of the corpus. The effect on the aggregate at the corpus's natural
mix is
\begin{equation}
\Delta_{\mathrm{nat}}=\pi_C\,\Delta_C+(1-\pi_C)\,\Delta_T .
\label{eq:agg}
\end{equation}
Copying the neighbour's label makes $\Delta_C>0>\Delta_T$, so $\Delta_{\mathrm{nat}}$ can be positive, zero or negative
while the model's behaviour has changed; $\pi_C$ is 0.135 (SGD), 0.099 (STAR) and 0.232 (AMI).

\paragraph{What counts as leakage.} The answer depends on the evaluation claim. In \emph{Setting A}, generalisation to
unseen dialogues, no gold label of the test dialogue may enter training or the labelled demonstrations; the lag-1 row
is leakage. In \emph{Setting B}, prospective evaluation within a dialogue, earlier acts are admissible only if they are
genuinely known before the prediction, as a system's own executed actions are and as human dialogue acts usually are
not. Unlabelled earlier text of the test dialogue is not gold-label leakage in either setting: it is ordinary history in
Setting B and, in Setting A, a transductive use of test-dialogue text that is admissible only where the benchmark
permits it. Repeating the previous act is a legitimate baseline in Setting B and only a reference in Setting A.
Appendix~\ref{app:settings} classifies the channels.

\paragraph{Dependent windows across a split.} Random splits of segmented time series place segments or records of
the same subject or recording on both sides, which inflates scores in activity recognition
\citep{hammerla2015stick,dehghani2019quantitative,tello2024good}, clinical prediction from wearable sensors
\citep{saeb2017need} and EEG classification \citep{brookshire2024data}; overlapping test windows are not independent
tests even under subject-disjoint splits \citep{shi2026windows}, leakage taxonomies list such non-independence
among standard pitfalls \citep{kapoor2023leakage}, and block or purged cross-validation are the general
remedies \citep{roberts2017cross,lopezdeprado2018advances}. Training--test overlap has also been measured in NLP benchmarks \citep{elangovan2021memorization} and in widely used open-domain dialogue datasets \citep{wen2022overlap}. In predictive process monitoring, prefixes of the same case
cross temporal and random splits \citep{weytjens2021creating}, and for next-activity prediction most test prefixes
duplicate training prefixes, which leads to overestimated performance \citep{pfeiffer2025learning}. In vision
benchmarks, leaked near-duplicates usually inflate scores but can lower them when their labels differ
\citep{ramos2025leakage}. Our problem differs in two ways: the leaked item is not the test example or a copy of it but its predecessor, whose label is the
correct answer only for continuations; and transitions are the majority in all three corpora, so the two subgroups pull
the aggregate in opposite directions and its net change can have either sign. Dependence alone is not leakage: $K$-fold
cross-validation on lagged data is valid for purely autoregressive models whose errors are serially uncorrelated
\citep{bergmeir2018note}, so whether shared rows change a score depends on how the learner uses them, which we examine
with three learners. Pre-training contamination, whose measured effect depends on its form and frequency, model scale
and later training \citep{jiang2024investigating,singh2024contamination,schaeffer2026quantifying}, cannot produce the
within-query contrasts we measure, although it may interact with them.

\paragraph{Retrieved demonstrations and label copying.} Similarity-retrieved demonstrations beat random ones
\citep{liu2022makes,rubin2022learning}, and with random demonstrations the correctness of their labels matters little
\citep{min2022rethinking}; but a demonstration that is near-identical to the test input is copied, label included, even
when the label is wrong \citep{lyu2023zicl}, and accuracy rises with the share of demonstrations that share the query's
label \citep{chen2026rethinking}. Demonstrated labels can override semantics \citep{liu2026fixation}, copying-related
neurons can be pruned \citep{ali2026mitigating}, and demonstration placement changes predictions
\citep{cobbina2025where}. These results establish the behaviour we rely on; they do not
say where it enters an evaluation; a recent survey of data effects in in-context learning does not treat
demonstrations drawn from the test item's own source \citep{schoch2025good}, and support-set leakage has been studied for relational foundation models
\citep{upendra2026support}. We add the case in which the near-duplicate is the test dialogue's own previous window,
whose label consistency with the query is not a property of the retriever but of the target: consistent for a
continuation, inconsistent for a transition.

\paragraph{Dialogue-act protocols and provenance filters.} \citet{tanaka2019dialogue} cut SwDA conversations into
overlapping five-utterance windows and split the windows at random for next-act prediction, giving the model gold
previous acts at test time by design; their system could not be re-evaluated, and we report only the exposure that
its protocol implies (Appendix~\ref{app:external}). Similarity-retrieved demonstrations are used for dialogue state
tracking \citep{hegde2025factors} and for annotating pedagogical dialogue acts, where the pipeline excludes the target
utterance's own session from retrieval to avoid leakage \citep{lee2026domain}; the filter is thus in use, and we
measure what an analogous filter removes and costs in our setting. Fine-grained dialogue-act classification remains
difficult for language models \citep{qamar2025dialogue}.

\section{Experimental Protocol}
\label{sec:protocol}

\begin{figure*}[t]
\centering
\includegraphics[width=\textwidth]{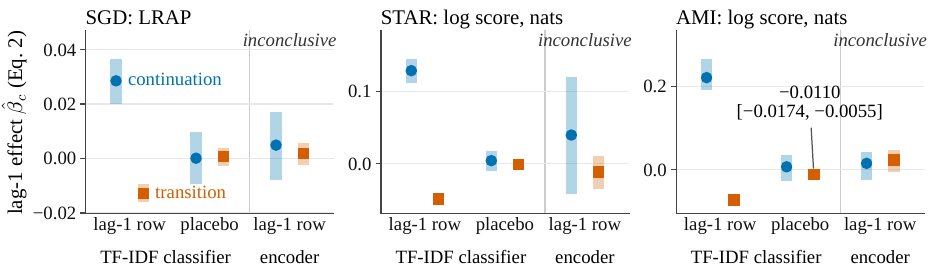}
\caption{Randomised presence design: effect $\hat\beta_c$ (Equation~\ref{eq:presence}) of the adjacent (lag-1) row
and of a non-overlapping placebo row on continuations (circles) and transitions (squares), with 95\% cluster-bootstrap
intervals (bands); the fine-tuned encoder is inconclusive in every corpus. Each panel has its own score; values in
Tables~\ref{tab:a-lexical}--\ref{tab:a-learners}.}
\label{fig:randomised}
\end{figure*}

\paragraph{Corpora.} We use SGD \citep{rastogi2020sgd}: 3,000 dialogues sampled from the training and development
portions, 30,501 system turns labelled with one of 13 distinct act sets, scored by label-ranking average precision
(LRAP); STAR \citep{mosig2020star}: 3,000 complete dialogues, wizard turns labelled with one of 161 actions; and the AMI
meeting corpus \citep{carletta2006ami}: one 200-segment excerpt of each of the 139 meetings, 27,744 segments labelled
with 15 dialogue-act types, organised in 36 participant-disjoint meeting series. STAR and AMI are scored by the log of
the normalised score of the gold class. All three official splits are dialogue- or meeting-disjoint; every exposure in
this paper is induced by our designs.

\paragraph{Randomised presence design.} For each corpus we run five repeats of five-fold cross-fitting over clusters.
In each repeat and evaluation fold, the targets of the evaluated dialogues are split at random into four groups. For
each group a \emph{grouped} fit trains on the other clusters only, and an \emph{exposed} fit replaces as many
label-matched training rows by the other groups' windows from the evaluated dialogues; both fits have the same size and
nearly the same label distribution and predict the same targets, so whether a target's lag-1 row is in its exposed
training set is randomised within dialogue. Its effect is estimated by a within-target regression. For target $t$ in
repeat $r$, the outcome is the exposed-minus-grouped score $D_{tr}$; the indicator $L_{tr}$ marks that the row of $p(t)$ is in the exposed training set, $L'_{tr}$ the same
for the next labelled target, and $N_{tr}$ counts the other same-dialogue rows present. Separately for each cell $c$ we fit
\begin{equation}
D_{tr}=\alpha_t+\beta_c L_{tr}+\delta_c L'_{tr}+\gamma_c N_{tr}+\varepsilon_{tr}
\label{eq:presence}
\end{equation}
with target fixed effects $\alpha_t$, over the targets whose lag-1 presence varies across repeats, and report the
\emph{lag-1 effect} $\hat\beta_c$. Controls: the same regression for a \emph{placebo} row, the nearest earlier
same-dialogue target more than five utterances back, whose window does not overlap $x_t$; separate fits (eight groups)
that \emph{purge} the two adjacent rows or as many label-matched non-adjacent rows; a permutation of the exposure rows'
labels within dialogue; and a naive row-level five-fold split. Learners: a TF-IDF (1--2-gram) logistic-regression
classifier (primary), a fine-tuned RoBERTa-base \citep{liu2019roberta} (secondary) and a TF-IDF $k$-nearest-neighbour
classifier (exploratory, added after the primary results). The prespecified decision required a positive continuation
effect, a positive continuation-minus-transition difference, an effect above the placebo and a positive purge contrast.

\begin{table*}[t]
\centering
\fontsize{10}{11}\selectfont
\setlength{\tabcolsep}{2.5pt}
\begin{tabular}{>{\raggedright\arraybackslash}p{3.7cm}ccc>{\raggedright\arraybackslash}p{3.7cm}}
\toprule
Control: estimand & SGD (LRAP) & STAR (nats) & AMI (nats) & What it shows \\
\midrule
Placebo row: lag-1 minus placebo effect, cont.$^{*}$ & \begin{tabular}[t]{@{}c@{}}$+0.028$\\{[}$0.017$, $0.040${]}\end{tabular} & \begin{tabular}[t]{@{}c@{}}$+0.125$\\{[}$0.107$, $0.142${]}\end{tabular} & \begin{tabular}[t]{@{}c@{}}$+0.214$\\{[}$0.192$, $0.240${]}\end{tabular} & Adjacency, not dialogue membership (cells: Fig.~3) \\
Purge of the adjacent minus as many matched rows, cont.$^{*}$ & \begin{tabular}[t]{@{}c@{}}$+0.020$\\{[}$0.017$, $0.023${]}\end{tabular} & \begin{tabular}[t]{@{}c@{}}$+0.133$\\{[}$0.123$, $0.143${]}\end{tabular} & \begin{tabular}[t]{@{}c@{}}$+0.202$\\{[}$0.191$, $0.214${]}\end{tabular} & The continuation gain is concentrated in the two adjacent rows \\
Labels permuted within dialogue: lag-1, cont. & \begin{tabular}[t]{@{}c@{}}$-0.008$\\{[}$-0.018$, $0.001${]}\end{tabular} & \begin{tabular}[t]{@{}c@{}}$+0.030$\\{[}$0.012$, $0.047${]}\end{tabular} & \begin{tabular}[t]{@{}c@{}}$-0.031$\\{[}$-0.052$, $-0.010${]}\end{tabular} & Needs the label; STAR keeps about a quarter \\
Lead-1 row: effect on transitions & \begin{tabular}[t]{@{}c@{}}$-0.004$\\{[}$-0.007$, $-0.001${]}\end{tabular} & \begin{tabular}[t]{@{}c@{}}$-0.038$\\{[}$-0.043$, $-0.032${]}\end{tabular} & \begin{tabular}[t]{@{}c@{}}$-0.005$\\{[}$-0.021$, $0.006${]}\end{tabular} & Predicted gain not observed \\
Fine-tuned encoder: lag-1, cont. & inconcl. & inconcl. & inconcl. & No detectable effect, not evidence of none (Fig.~3; Table~5) \\
TF-IDF $k$-NN (exploratory): lag-1, cont. & \begin{tabular}[t]{@{}c@{}}$+0.201$\\{[}$0.186$, $0.217${]}\end{tabular} & \begin{tabular}[t]{@{}c@{}}$+0.773$\\{[}$0.642$, $0.907${]}\end{tabular} & \begin{tabular}[t]{@{}c@{}}$+0.999$\\{[}$0.868$, $1.173${]}\end{tabular} & Largest SGD effect; added after primary results \\
\bottomrule
\end{tabular}
\caption{Controls of the randomised presence design (TF-IDF classifier unless stated), 95\% cluster-bootstrap intervals; cont.: continuation targets; $^{*}$prespecified decision criterion. All estimands: Tables~\ref{tab:a-lexical}--\ref{tab:a-learners}.}
\label{tab:controls}
\end{table*}

\paragraph{Retrieval in-context learning.} For SGD and AMI we draw 750 queries per corpus and sample (300
continuations and 450 transitions; at most one per SGD dialogue). For query $q$ the \emph{filtered} pool holds every
target outside $q$'s cluster (its dialogue, near-duplicate group or AMI meeting series), and the \emph{exposed} pool
adds the earlier labelled targets of $q$'s own dialogue. The four pool members most similar to $q$ under TF-IDF cosine,
each a window with its gold act, form the demonstrations $D^E_q$ or $D^F_q$, shown in ascending similarity. A model
scores every string of the fixed label inventory (13 SGD act sets, 15 AMI types) by the summed log probability of its
tokens, without length normalisation; $s(q;D)$ is the log of the gold string's probability renormalised over the
inventory, and $h(q;D)$ indicates that the gold string ranks first. The cell effect over the queries $Q_c$ of cell $c$ is
\begin{equation}
\Delta^{s}_c=\frac{1}{|Q_c|}\sum_{q\in Q_c}\bigl[s(q;D^E_q)-s(q;D^F_q)\bigr]
\label{eq:icl}
\end{equation}
in nats; $\Delta^{h}_c$ is the same with $h$, in top-1 accuracy. The natural-weight aggregate reweights the stratified
queries to the corpus's $\pi_C$, so it equals Equation~\eqref{eq:agg} applied to $\Delta^{h}_C$ and $\Delta^{h}_T$.
Further conditions replace each same-dialogue demonstration by the most similar other-dialogue window with the same or
with a different label, use random demonstrations, or use none. Models are Qwen2.5-7B-Instruct \citep{qwen2024qwen25}
and Mistral-7B-Instruct-v0.3 \citep{jiang2023mistral}. A fresh sample of 750 queries per corpus that the first never
used was scored after the first analysis; for SGD it comes from other dialogues, for AMI from the same 36 series, so it
tests target sampling, not new participants. An exploratory STAR arm used Qwen only. Decision rules for both samples
were fixed before scoring.

\paragraph{Statistics.} Intervals are 95\% cluster-bootstrap percentile intervals with 2,000 replicates over
dialogues (SGD, where no near-duplicates were found), near-duplicate dialogue groups (STAR) or meeting series (AMI);
each replicate reweights the targets or queries of the drawn clusters, for Equation~\eqref{eq:presence} inside its
normal equations. Aggregate intervals are computed directly, not from cell intervals. Scores are never compared across
corpora.

\section{Results}
\label{sec:results}

\subsection{The adjacent labelled row drives a cell asymmetry}

For the lexical classifier, the lag-1 coefficient $\hat\beta_c$ of Equation~\eqref{eq:presence} is positive on
continuations and negative on transitions in all three corpora (Figure~\ref{fig:randomised}, Table~\ref{tab:controls}):
$+0.0285$ [$+0.0201$, $+0.0364$] and $-0.0129$ [$-0.0161$, $-0.0095$] LRAP on SGD, and $+0.2209$ [$+0.1910$, $+0.2648$]
and $-0.0722$ [$-0.0781$, $-0.0660$] nats on AMI. The prespecified decision was met in all three corpora, and the
continuation gain from exposure was positive in all five repeats. The effect is specific to the adjacent row: a non-overlapping placebo row from the same dialogue has no detectable effect on continuations but shows a small negative effect on AMI transitions, $-0.0110$ [$-0.0174$, $-0.0055$] nats (Figure~\ref{fig:randomised}). In separate fits, purging the two adjacent rows removes 84--93\% of the continuation gain, whereas purging as many label-matched other rows removes $-2$\% to $+1$\%. It depends on the neighbour's label:
permuting the exposure rows' labels within dialogue leaves no positive effect on SGD and AMI and about a quarter on
STAR, whose templated wizard turns often make a permuted label coincide with the continuation label. The predicted
benefit of the following row on transitions was not observed; copying its label helps only the 7--20\% of transitions
whose next act equals theirs. On AMI, the prespecified lexical-classifier decision also holds after excluding 13,193 targets whose history contains overlapping-onset speech (Appendix~\ref{app:randomised}).

The size of the effect depends on the learner (Table~\ref{tab:a-learners}): on the same targets (exploratory;
repeats 0--2), the lexical classifier's lag-1 continuation effect exceeds the fine-tuned encoder's by $+0.0282$ [$+0.0113$, $+0.0443$] LRAP on SGD,
$+0.0844$ [$+0.0004$, $+0.1705$] nats on STAR (narrowly) and $+0.2026$ [$+0.1432$, $+0.2977$] on AMI. The encoder alone shows no detectable adjacency effect (the SGD encoder is near ceiling) but gains
diffusely on AMI; the exploratory $k$-nearest-neighbour learner has the largest effect on the bounded LRAP of SGD,
$+0.2009$; its unbounded log scores are not comparable. Which property of a learner makes it
copy is not identified.

\begin{figure}[t]
\centering
\includegraphics[width=\columnwidth]{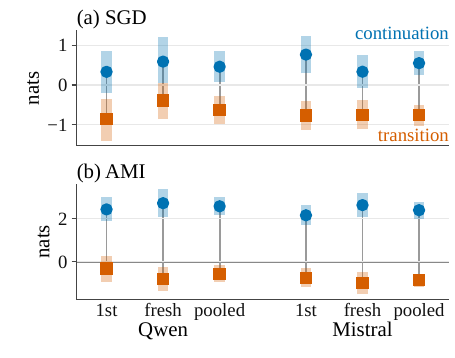}
\caption{Exposed minus filtered pool (Equation~\ref{eq:icl}): gold-act log probability (nats) on continuations
(circles) and transitions (squares), joined per sample (300 and 450 queries; pooled: both); bands: 95\% intervals.}
\label{fig:llm}
\end{figure}

\begin{figure}[t]
\centering
\includegraphics[width=\columnwidth]{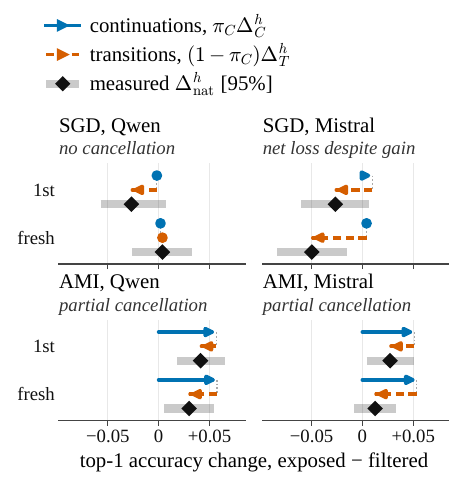}
\caption{Natural-weight top-1 change, exposed minus filtered (Equation~\ref{eq:agg}): from 0, the weighted continuation
contribution (blue arrow), then the weighted transition contribution (orange arrow), ending at the measured change
(diamond, its 95\% interval as a band); $\pi_C=0.135$ (SGD), $0.232$ (AMI). Cell intervals: Table~\ref{tab:a-llm}.}
\label{fig:masking}
\end{figure}

\begin{table}[t]
\centering
\fontsize{10}{11}\selectfont
\setlength{\tabcolsep}{3pt}
\begin{tabular}{llrll}
\toprule
 & & \multicolumn{2}{c}{Interaction (nats)} & Prespecified \\ \cmidrule(lr){3-4}
Arm & Sample & est. & 95\% CI & decision \\
\midrule
SGD & 1st & $+1.19$ & [$0.48$, $1.92$] & \textbf{partial} \\
Qwen & fresh & $+0.98$ & [$0.25$, $1.74$] & \textbf{partial} \\
 & pooled & $+1.09$ & [$0.60$, $1.63$] & -- \\
\midrule
SGD & 1st & $+1.53$ & [$0.92$, $2.11$] & present \\
Mistral & fresh & $+1.08$ & [$0.55$, $1.64$] & replicated \\
 & pooled & $+1.31$ & [$0.93$, $1.74$] & -- \\
\midrule
AMI & 1st & $+2.74$ & [$1.99$, $3.47$] & present \\
Qwen & fresh & $+3.51$ & [$2.58$, $4.41$] & replicated \\
 & pooled & $+3.13$ & [$2.54$, $3.70$] & -- \\
\midrule
AMI & 1st & $+2.89$ & [$2.26$, $3.54$] & present \\
Mistral & fresh & $+3.62$ & [$2.81$, $4.36$] & replicated \\
 & pooled & $+3.26$ & [$2.76$, $3.72$] & -- \\
\midrule
STAR & 1st & $+1.28$ & [$0.71$, $1.81$] & expl. \\
Qwen &  &  &  &  \\
\bottomrule
\end{tabular}
\caption{Interaction, continuation minus transition effect, and the decision fixed before scoring: \emph{present} (gain and interaction), \emph{replicated} (every Qwen first-sample finding), \textbf{partial} (a component missed); expl.: exploratory arm. AMI samples share all 36 series; all cells in Table~\ref{tab:a-llm}.}
\label{tab:llm}
\end{table}

\subsection{Retrieved demonstrations reproduce the asymmetry}

Figure~\ref{fig:llm} shows the paired contrast of Equation~\eqref{eq:icl}. As point estimates, exposure raises the gold
act's log score on continuations and lowers it on transitions in all eight model--corpus--sample arms. Pooled over both
samples (600 continuations and 900 transitions), the continuation gain is $+0.46$ [$+0.08$, $+0.85$] nats for Qwen and
$+0.55$ [$+0.25$, $+0.86$] for Mistral on SGD, and $+2.58$ [$+2.17$, $+3.01$] and $+2.40$ [$+1.99$, $+2.79$] on AMI; the
transition losses are $-0.63$ [$-0.98$, $-0.29$], $-0.76$ [$-1.02$, $-0.51$], $-0.55$ [$-0.93$, $-0.13$] and $-0.86$
[$-1.15$, $-0.57$]. The prespecified continuation--transition interaction excludes zero in all eight arms
(Table~\ref{tab:llm}), but four of the sixteen single-cell intervals include zero. For SGD/Qwen under the prespecified primary order (order A), the first-sample continuation gain was $+0.33$ [$-0.20$, $+0.87$], and the fresh-sample transition loss was $-0.39$ [$-0.86$, $+0.06$]. Both intervals include zero, so the corresponding decisions remain \emph{partial} (Table~\ref{tab:a-llm}). Under the alternative seeded slot order (order B), the fresh-sample transition loss was $-1.00$ [$-1.50$, $-0.53$]. This sensitivity result makes the transition loss detectable under order B but does not change the primary-order replication decision.
Mistral on SGD and both AMI arms met their decisions in both samples, although Mistral's fresh SGD continuation gain,
not a replication target, missed. The AMI samples share all 36 meeting series, so the AMI replication speaks to target
sampling, not to new participants. The exploratory STAR arm shows a continuation gain without a detectable transition
loss. Differences between the samples cover zero for every prespecified estimand (median half-widths 0.55--0.78 nats),
which is not evidence of equivalence (Appendix~\ref{app:llm}).

\begin{figure}[t]
\centering
\includegraphics[width=\columnwidth]{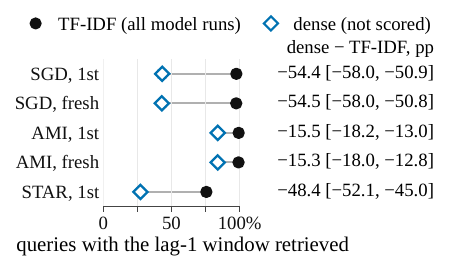}
\caption{Previous-act copying on transition queries, where predicting the previous act is always incorrect. The left panel shows copy rates under zero-shot, filtered, and exposed retrieval; the right panel shows the paired exposed-minus-filtered difference with its 95\% cluster-bootstrap interval. Each sample contains 450 transition queries. Numerical results are reported in Table~\ref{tab:a-copy}.}
\label{fig:prevalence}
\end{figure}

\begin{figure}[t]
\centering
\includegraphics[width=\columnwidth]{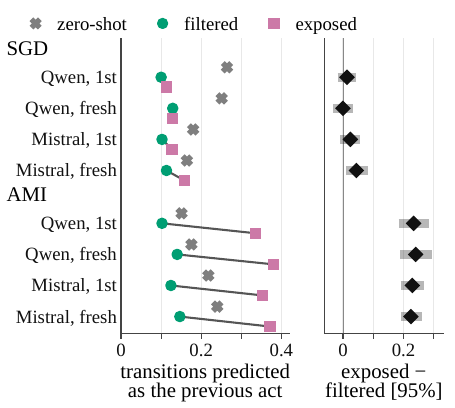}
\caption{Previous-act predictions: share of the 450 transition queries per sample predicted as the previous act (always wrong)
with no demonstrations, the filtered pool and the exposed pool, top to bottom in each row (the offsets carry no data; a
line joins filtered and exposed); right: exposed minus filtered with its 95\% interval.}
\label{fig:copy}
\end{figure}

\subsection{Aggregate scores hide the response}

Equation~\eqref{eq:agg} allows an aggregate to move in either direction, and it does. Under a naive row-level split,
the lexical classifier's change over a grouped split, on all targets including dialogue-initial ones (which
Equation~\eqref{eq:agg} does not cover), is \emph{negative} on SGD, $-0.0094$ [$-0.0104$, $-0.0085$] LRAP, and STAR,
$-0.0312$ [$-0.0346$, $-0.0281$] nats, and positive on AMI, $+0.0411$ [$+0.0325$, $+0.0499$], while continuation and
transition cells move in opposite directions in every corpus (Appendix~\ref{app:randomised}). The $k$-nearest-neighbour
learner's aggregate falls under the row split on SGD and STAR, so a strong leak can look like a deficit. For the
language models, where Equation~\eqref{eq:agg} holds on one score and one query set (Figure~\ref{fig:masking}), the
natural-weight change in top-1 accuracy pooled over both samples is $-0.011$ [$-0.033$, $+0.011$] for Qwen and $-0.038$
[$-0.062$, $-0.013$] for Mistral on SGD, and $+0.036$ [$+0.017$, $+0.055$] and $+0.020$ [$+0.004$, $+0.035$] on AMI,
although AMI continuation accuracy improves under exposure (Figure~\ref{fig:masking}). The weighted contributions cancel partly on AMI,
where the fresh-sample Mistral interval includes zero; on SGD, Qwen shows no cancellation because neither top-1 cell
moves, and for Mistral the transition loss outweighs a small continuation gain. Comparing aggregates with and without
exposure is therefore not a leakage test.

\subsection{How often retrieval exposes the test dialogue}

With the TF-IDF pipeline used throughout, the test dialogue's previous window is among the four retrieved demonstrations
for 97.6\% of SGD and 99.3\% of AMI queries in the first sample, usually in the slot nearest the query, but for only
75.6\% on STAR, whose templated wizard turns make other dialogues nearly as similar (Figure~\ref{fig:prevalence}, Table~\ref{tab:a-prevalence}). On
the same queries and pool, a second pipeline that embeds the rendered window text with a pretrained dense encoder
\citep{xiao2024cpack} retrieves it for 83.9\% of AMI queries but only 43.2\% of SGD and 27.2\% of STAR queries; every
paired difference is negative, and no dense-only lag-1 retrieval was observed in the evaluated SGD or AMI samples (Appendix~\ref{app:prevalence}). The two pipelines differ in representation as well as model, so this is a comparison of two concrete pipelines, not
of dense and lexical retrieval in general, and no language model was run on dense-retrieved demonstrations.

\subsection{Diagnostic comparisons and filtering}
\label{sec:diagnostics}

\paragraph{Diagnostic comparisons.} For a trained model on a released row-level split, comparing test targets whose lag-1 row fell
into the training fold with those whose lag-1 row did not, separately per cell, is observational rather than randomised,
yet from the row-split test scores alone it agrees with the randomised effect in sign and approximate size: on
continuations $+0.0316$~[$+0.0207$,~$+0.0421$] (SGD), $+0.0898$ [$+0.0374$, $+0.1435$] (STAR) and $+0.2247$ [$+0.1989$,
$+0.2519$] (AMI), against randomised effects of $+0.0285$, $+0.1286$ and $+0.2209$. For a retrieval pipeline, the share
of the 450 transition queries predicted as the previous act (always wrong) needs a reference, an inference-only run with
the filtered pool or the zero-shot rate. On AMI, exposure raises it from 0.102 to 0.336 and
from 0.140 to 0.380 for Qwen (first and fresh sample) and from 0.124 to 0.353 and 0.147 to 0.371 for Mistral, above the
zero-shot rates (Figure~\ref{fig:copy}, Table~\ref{tab:a-copy}); on SGD it changes by at most 0.05 (detectably only for Mistral, fresh sample)
and stays below the zero-shot rate. A split of exposed scores by cell is not
by itself a leak detector: high continuation and low transition accuracy also describe an honest model on a persistent
corpus.

\paragraph{Cost of the filter.} On SGD, filtered retrieval is useful: Qwen reaches 0.516 and 0.515 top-1 accuracy in the
two samples, against 0.261 and 0.289 zero-shot and 0.346 and 0.361 with random demonstrations (Table~\ref{tab:a-filter},
Figure~\ref{fig:utility}), and filtering lowered accuracy detectably in no SGD arm (Figure~\ref{fig:masking}), which
does not make it free. On AMI, filtered retrieval is not detectably different from zero-shot for Qwen and falls below it
for Mistral, by $-0.097$ [$-0.131$, $-0.065$] and $-0.074$ [$-0.101$, $-0.046$]; every retrieval condition of both
models scores below repeating the previous act, which scores the continuation share of 0.232, while Mistral's zero-shot
accuracy is at that level. The small aggregate gain from exposure on AMI comes from continuations and is partly paid for
by transitions, a move toward this persistence baseline. In Setting B that baseline is legitimate and these models do
not beat it; in Setting A, the test dialogue's gold previous act is unavailable, and the baseline only shows what the
exposed pool lets a model imitate. For Qwen on AMI, a weaker filter that excludes only the query meeting but permits other meetings of the same participants showed no detectable change in natural-weight top-1 accuracy relative to the stricter series-disjoint filter in either sample. This does not establish equivalence between the filters.

\subsection{What the leaked window carries}

Placing the previous act's label on the most similar other-dialogue window, instead of a different label on such a
window, raises continuation scores in all eight arms, so label placement alone reproduces the continuation gain. On SGD
the same-dialogue window is \emph{less} useful than that replacement (detectably in three of four arms), while on AMI it
adds 1.45 to 2.60 nats more. This AMI increment has no common support in similarity: none of 598 pooled continuation
queries has a same-dialogue window within 0.1 cosine of its replacement, and the increment grows with the gap
(Appendix~\ref{app:decomposition}). It is the contrast that natural retrieval produces, not a provenance effect at fixed
similarity.

An exploratory control showed the same-dialogue window \emph{without any label}: the unlabelled window raised
continuation scores beyond a length-matched unlabelled window from another dialogue in all four model--corpus arms, by
$+0.43$ [$+0.12$, $+0.77$] to $+1.14$ [$+0.64$, $+1.67$] nats, whereas a wrong label on the same text was harmful in
three of the four arms. The other window was far less similar, so this is not a contrast at equal similarity. A
follow-up with one model split the unlabelled window into the utterances already inside the query window and the older
ones. On AMI the shared utterances carried the gain, $+0.83$ [$+0.16$, $+1.54$] nats over a token-matched excerpt from
another meeting, with no detectable effect of the older utterance; on SGD neither part showed a detectable effect,
$+0.23$ [$-0.15$, $+0.59$] for the shared part. Descriptively, the AMI gain concentrated where the shared utterances
carry speaker tags relative to a different participant, so repetition of shared text and its re-presentation under
another speaker's frame cannot be separated. The follow-up detected no shift toward the previous act, so an earlier
reading of the text effect as a persistence cue is not confirmed (Appendix~\ref{app:unlabelled}). In Setting A only the
label channel is gold-label leakage; the unlabelled text is earlier history.

\section{Discussion}
\label{sec:discussion}

In both designs, exposure to the labelled predecessor helped continuations and hurt transitions for the lexical
classifier and both language models, by amounts that depended on learner and corpus, so a natural aggregate rose, fell
or stayed flat and cannot certify a leak-free evaluation. All exposure was induced; no published system was measured.

What counts as leakage depends on the claim. For unseen dialogues (Setting A), splitting by dialogue or near-duplicate
cluster before windowing and excluding the test dialogue's cluster from demonstration pools keeps its gold labels out;
shared text alone is not label leakage. For prospective use (Setting B), earlier acts are admissible only if observed at
prediction time, and repeating the previous act is the baseline to beat. Reporting both cells beside the aggregate makes
Equation~\eqref{eq:agg} visible; the two diagnostics require row-split predictions or an inference-only reference, and an
undetected filter cost is not equivalence. Masking and wrong labels are not neutral
controls (Table~\ref{tab:a-negative}). Finally, label strings were scored by summed token log probabilities without
length normalisation; scoring one inventory under both pools does not make length effects cancel, so the language-model
contrasts may depend on this untested choice.

\section{Conclusion}

When a test target's labelled predecessor reached training or retrieval, continuation scores rose and transition scores
fell: in a randomised design for a lexical classifier on SGD, STAR and AMI (not detectably for a fine-tuned encoder),
and with TF-IDF-retrieved demonstrations for two 7B language models on SGD and AMI in two samples (partially for one
model on SGD). Natural aggregates rose, fell or stayed flat. A dense pipeline retrieved the predecessor less often; no
model was scored on its demonstrations. Cell-wise reporting, provenance filtering for unseen-dialogue
claims, and the two diagnostics make the problem visible at little cost.

\section*{Limitations}

All exposure was induced on corpora with clean official splits; we make no claim about any published result. A planned
independent evaluation of the closest published row-split next-act system \citep{tanaka2019dialogue} was not feasible because its
implementation, split indices and tag mapping are not available and no licensed copy of its corpus was at hand; we
report only exposure rates implied by its published protocol (Appendix~\ref{app:external}). The language-model results
come from two 7B instruction-tuned models, one prompt format, four demonstrations and TF-IDF retrieval; whether models
react as strongly to dense-retrieved same-dialogue demonstrations was not tested, and the follow-up decomposition used
one model only. On AMI, top-1 accuracy under random, filtered, or exposed demonstrations was low (0.090--0.163) and below the previous-act reference accuracy of 0.232. Weak task competence may therefore amplify sensitivity to copied labels or repeated text, and the magnitude of these effects should not be assumed to generalize to stronger models. No run used a model's own earlier predictions in place of gold labels, as Setting B requires where
acts are not observed. The prevalence comparison covers two
concrete pipelines that differ in representation and model; a fine-tuned or dialogue-specific retriever may behave
differently. The SGD language-model components are individually modest and each missed in at least one arm; both SGD
decisions for Qwen are partial. AMI inference rests on 36 meeting series shared by all samples, so the AMI results do
not generalise to new participants. The same-meeting increment is not identified at fixed similarity, and the
unlabelled-text effect cannot be separated from near-duplication or a change of speaker frame. Pre-training
contamination of public corpora cannot be excluded; it cannot create a within-query contrast between pools but may
interact with one, so absolute accuracies are not capability estimates. The corpora are English task-oriented and meeting
dialogues; counselling and clinical corpora, where split units were most often unspecified in our review, are not
represented. We identify behavioural dependencies only, not mechanisms inside the models.

\section*{Ethical Considerations}

SGD (simulator-generated dialogue outlines paraphrased by crowd workers; CC BY-SA 4.0), STAR (Wizard-of-Oz dialogues
between crowd workers; MIT) and AMI (scenario-based and other recorded meetings; manual annotations under CC BY 4.0) are
public research corpora used under their licences; no clinical or private data were used. The paper reports only
aggregate estimates with their intervals; no participant or dialogue identifier, utterance, prompt or per-example score
is reported or released. The work concerns evaluation methodology, evaluates no clinical system and validates no model
for deployment. Exposure to the neighbour's label can make a system look better or worse than it is; reporting
continuation and transition cells exposes it. AI-based language tools assisted with manuscript preparation. The authors take full responsibility for the manuscript content, code, and data handling.

\bibliography{references}

\appendix

\makeatletter\setlength{\@dblfptop}{0pt}\setlength{\@dblfpsep}{\dblfloatsep}\setlength{\@dblfpbot}{0pt plus 1fil}\makeatother

\section{Evaluation Settings and Leakage Channels}
\label{app:settings}

Table~\ref{tab:a-channels} classifies the channels through which material from the test dialogue can reach a windowed
next-act model. Under Setting A (generalisation to unseen dialogues), the gold acts of earlier windows are leakage
whenever they enter training or the labelled demonstrations. Under Setting B (prospective evaluation within a dialogue),
an earlier act is admissible only if it is genuinely observed at run time: a dialogue system knows its own executed
actions (SGD system acts come from a dialogue simulator; STAR wizards mostly pick their reply, and with it its action
label, from a ranked list of suggested responses), whereas human dialogue acts
such as those of AMI are observed only if someone annotates them online. The admissible substitute is then the model's
own earlier prediction, not the gold label.

\begin{table*}[t]
\centering
\footnotesize
\begin{tabular}{*{4}{>{\raggedright\arraybackslash}p{3.45cm}}}
\toprule
Channel & Example in this paper & Setting A: unseen dialogues & Setting B: prospective, observed history \\
\midrule
Later windows of the test dialogue (may contain the target utterance) & pool allowing later windows & leakage & leakage (the future is not observed) \\
Earlier windows of the test dialogue with their gold acts & exposed pool; row-level split & leakage & admissible only for acts genuinely observed at run time \\
Earlier windows of the test dialogue without labels & unlabelled-window control & not label leakage; transductive use of test-dialogue text, admissible only if the benchmark permits it & admissible history \\
Near-duplicate dialogues across the split & SGD and STAR near-duplicate groups & leakage unless clustered before splitting & ordinary training data \\
Same participants, another meeting & AMI series-level pool & definition-dependent (no measurable effect here) & admissible \\
Pre-training contamination & both language models postdate the corpora & shifts absolute levels; cannot create a within-query pool contrast & same \\
\bottomrule
\end{tabular}
\caption{Leakage channels for windowed dialogue data by evaluation claim.}
\label{tab:a-channels}
\end{table*}

\section{Split Practice in Published Work}
\label{app:audit}

The prevalence figures reported in the Introduction are based on a review of 78 context-window dialogue studies published between 2019 and 2026. Each study was examined using its full text and, where available, released code to identify the unit of data splitting. Four studies (5.1\%, Wilson 95\% interval 2.0--12.5\%) reported a confirmed row- or window-level split, including one verified at the code level. Another 22 studies (28.2\%) did not clearly specify their splitting unit, 46 used dialogue-disjoint or officially grouped splits, and 6 employed mixed splitting strategies across components.

Two authors independently coded all 78 studies using the same coding criteria. Disagreements between the two sets of classifications were resolved through discussion and consensus. The reported proportions describe the reviewed sample rather than the broader research literature, and incomplete coverage of eligible studies may introduce selection bias.

Within the reviewed sample, unspecified splitting units were concentrated in counselling and medical corpora. Importantly, identifying a row-level split does not establish that the corresponding published results are inflated; it only indicates a condition under which the exposure investigated in this study may occur.

\section{Randomised Presence Design: Complete Results}
\label{app:randomised}

The design holds training-set size and label distribution fixed while randomising, within each evaluated dialogue, which
neighbouring windows are present. Table~\ref{tab:a-lexical} lists every estimand for the primary classifier and
Table~\ref{tab:a-learners} lists those of the secondary and exploratory learners. Masking the target text is not a neutral
control: relative to a placebo mask it lowers the continuation gain in all three corpora, while most of the gain
survives it. On AMI, the prespecified decision was also met after removing the 13,193 targets with an overlapping-onset
history segment. Both neighbours pull transition
predictions toward their own labels, while the placebo row does not.

\paragraph{Estimation details.} Equation~\eqref{eq:presence} is fitted by least squares after demeaning each target's
five repeats. Targets without a lag-1 row, or whose lag-1 presence does not vary across the repeats, drop out
(Table~\ref{tab:a-lexical} gives the number that remain); a missing lead-1 row counts as absent, and $N_{tr}$ excludes
the two neighbours. The interaction is the difference of the two cell coefficients. In each of the 2,000 bootstrap
replicates, clusters are drawn with replacement and each target's cross-products over the repeats enter the normal
equations with its cluster's multiplicity; the 95\% interval is the 2.5th to 97.5th percentile. The purge contrasts come
from separate fits with eight target groups per evaluation fold; the purge share is the ratio of the repeat-averaged
cell mean of the exposed-minus-purged score to that of the exposed-minus-grouped score. The row-split diagnostic of
Section~\ref{sec:diagnostics} compares, within a cell, the row-split scores of targets whose lag-1 row fell into the
training folds with those whose lag-1 row did not. Under a uniformly random row split this placement is itself random,
but the comparison is between targets, unadjusted and confined to one split, so it is a descriptive diagnostic, not an
estimate of the adjusted effect in Equation~\eqref{eq:presence}.

\begin{table*}[t]
\centering
\footnotesize
\setlength{\tabcolsep}{2.2pt}
\begin{tabular}{>{\raggedright\arraybackslash}p{3.7cm}rrrrrrrrr}
\toprule
 & \multicolumn{3}{c}{SGD (LRAP)} & \multicolumn{3}{c}{STAR (nats)} & \multicolumn{3}{c}{AMI (nats)} \\
\cmidrule(lr){2-4} \cmidrule(lr){5-7} \cmidrule(lr){8-10}
Estimand (TF-IDF classifier) & est. & lo & hi & est. & lo & hi & est. & lo & hi \\
\midrule
Lag-1 present, continuation & $+0.0285$ & $0.0201$ & $0.0364$ & $+0.1286$ & $0.1122$ & $0.1443$ & $+0.2209$ & $0.1910$ & $0.2648$ \\
Lag-1 present, transition & $-0.0129$ & $-0.0161$ & $-0.0095$ & $-0.0492$ & $-0.0536$ & $-0.0446$ & $-0.0722$ & $-0.0781$ & $-0.0660$ \\
Lag-1, continuation minus transition & $+0.0413$ & $0.0325$ & $0.0503$ & $+0.1778$ & $0.1611$ & $0.1941$ & $+0.2931$ & $0.2625$ & $0.3379$ \\
Placebo row present, continuation & $+0.0001$ & $-0.0093$ & $0.0097$ & $+0.0041$ & $-0.0099$ & $0.0177$ & $+0.0070$ & $-0.0265$ & $0.0362$ \\
Lag-1 minus placebo, continuation & $+0.0284$ & $0.0168$ & $0.0396$ & $+0.1246$ & $0.1068$ & $0.1422$ & $+0.2140$ & $0.1923$ & $0.2397$ \\
Lead-1 present, transition & $-0.0039$ & $-0.0074$ & $-0.0006$ & $-0.0376$ & $-0.0426$ & $-0.0324$ & $-0.0051$ & $-0.0209$ & $0.0056$ \\
Adjacent purge minus placebo purge, continuation & $+0.0200$ & $0.0170$ & $0.0231$ & $+0.1333$ & $0.1235$ & $0.1431$ & $+0.2024$ & $0.1914$ & $0.2145$ \\
Lag-1, labels permuted within dialogue & $-0.0084$ & $-0.0180$ & $0.0010$ & $+0.0296$ & $0.0124$ & $0.0470$ & $-0.0306$ & $-0.0521$ & $-0.0096$ \\
Row split minus grouped split, all targets & $-0.0094$ & $-0.0104$ & $-0.0085$ & $-0.0312$ & $-0.0346$ & $-0.0281$ & $+0.0411$ & $0.0325$ & $0.0499$ \\
Row split minus grouped split, continuation & $+0.0175$ & $0.0147$ & $0.0205$ & $+0.1406$ & $0.1291$ & $0.1516$ & $+0.2256$ & $0.2072$ & $0.2444$ \\
Row split minus grouped split, transition & $-0.0148$ & $-0.0159$ & $-0.0137$ & $-0.0513$ & $-0.0542$ & $-0.0482$ & $-0.0147$ & $-0.0199$ & $-0.0096$ \\
Diagnostic from row-split scores, continuation & $+0.0316$ & $0.0207$ & $0.0421$ & $+0.0898$ & $0.0374$ & $0.1435$ & $+0.2247$ & $0.1989$ & $0.2519$ \\
Diagnostic from row-split scores, transition & $-0.0131$ & $-0.0173$ & $-0.0088$ & $-0.0482$ & $-0.0690$ & $-0.0279$ & $-0.0493$ & $-0.0629$ & $-0.0338$ \\
Share of the continuation gain removed by the adjacent purge & $+0.93$ & $0.85$ & $1.02$ & $+0.84$ & $0.81$ & $0.86$ & $+0.84$ & $0.80$ & $0.88$ \\
Share of the continuation gain removed by the placebo purge & $-0.02$ & $-0.08$ & $0.04$ & $0.00$ & $-0.02$ & $0.01$ & $+0.01$ & $0.00$ & $0.02$ \\
Target text masked minus placebo mask, continuation gain & $-0.0099$ & $-0.0119$ & $-0.0078$ & $-0.0060$ & $-0.0082$ & $-0.0037$ & $-0.0233$ & $-0.0266$ & $-0.0205$ \\
Targets with varying lag-1 presence ($n$) & \multicolumn{3}{c}{2{,}310} & \multicolumn{3}{c}{1{,}099} & \multicolumn{3}{c}{4{,}865} \\
Prespecified decision & \multicolumn{3}{c}{supported} & \multicolumn{3}{c}{supported} & \multicolumn{3}{c}{supported} \\
\bottomrule
\end{tabular}
\caption{Complete estimands of the randomised presence design for the TF-IDF classifier. The row-split contrasts compare a naive 5-fold row-level split with the grouped split on the same targets; the diagnostic contrasts test targets whose lag-1 row fell into the training fold with those whose lag-1 row did not, using only row-split scores.}
\label{tab:a-lexical}
\end{table*}

\begin{table*}[t]
\centering
\footnotesize
\setlength{\tabcolsep}{3pt}
\begin{tabular}{>{\raggedright\arraybackslash}p{4.0cm}ccc}
\toprule
Estimand & SGD (LRAP) & STAR (nats) & AMI (nats) \\
\midrule
\emph{Fine-tuned RoBERTa} &  &  &  \\
Lag-1 present, continuation & \begin{tabular}[t]{@{}c@{}}$+0.0049$\\{[}$-0.0079$, $0.0172${]}\end{tabular} & \begin{tabular}[t]{@{}c@{}}$+0.0395$\\{[}$-0.0415$, $0.1199${]}\end{tabular} & \begin{tabular}[t]{@{}c@{}}$+0.0153$\\{[}$-0.0247$, $0.0418${]}\end{tabular} \\
Lag-1 present, transition & \begin{tabular}[t]{@{}c@{}}$+0.0018$\\{[}$-0.0024$, $0.0058${]}\end{tabular} & \begin{tabular}[t]{@{}c@{}}$-0.0119$\\{[}$-0.0353$, $0.0105${]}\end{tabular} & \begin{tabular}[t]{@{}c@{}}$+0.0231$\\{[}$-0.0063$, $0.0484${]}\end{tabular} \\
Lag-1 minus placebo, continuation & \begin{tabular}[t]{@{}c@{}}$+0.0107$\\{[}$-0.0091$, $0.0301${]}\end{tabular} & \begin{tabular}[t]{@{}c@{}}$+0.0745$\\{[}$-0.0164$, $0.1629${]}\end{tabular} & \begin{tabular}[t]{@{}c@{}}$+0.0221$\\{[}$-0.0109$, $0.0498${]}\end{tabular} \\
Any same-dialogue exposure, continuation & \begin{tabular}[t]{@{}c@{}}$+0.0007$\\{[}$-0.0018$, $0.0030${]}\end{tabular} & \begin{tabular}[t]{@{}c@{}}$+0.0143$\\{[}$0.0002$, $0.0286${]}\end{tabular} & \begin{tabular}[t]{@{}c@{}}$+0.0448$\\{[}$0.0352$, $0.0538${]}\end{tabular} \\
Any same-dialogue exposure, transition & \begin{tabular}[t]{@{}c@{}}$-0.0003$\\{[}$-0.0011$, $0.0005${]}\end{tabular} & \begin{tabular}[t]{@{}c@{}}$+0.0007$\\{[}$-0.0036$, $0.0049${]}\end{tabular} & \begin{tabular}[t]{@{}c@{}}$+0.0046$\\{[}$0.0019$, $0.0078${]}\end{tabular} \\
Row split minus grouped split, all targets & \begin{tabular}[t]{@{}c@{}}$-0.0010$\\{[}$-0.0022$, $0.0001${]}\end{tabular} & \begin{tabular}[t]{@{}c@{}}$+0.0050$\\{[}$-0.0012$, $0.0112${]}\end{tabular} & \begin{tabular}[t]{@{}c@{}}$+0.0190$\\{[}$0.0157$, $0.0223${]}\end{tabular} \\
Prespecified decision & inconcl. & inconcl. & inconcl. \\
\emph{TF-IDF $k$-NN ($k=10$, exploratory)} &  &  &  \\
Lag-1 present, continuation & \begin{tabular}[t]{@{}c@{}}$+0.2009$\\{[}$0.1857$, $0.2168${]}\end{tabular} & \begin{tabular}[t]{@{}c@{}}$+0.7727$\\{[}$0.6419$, $0.9071${]}\end{tabular} & \begin{tabular}[t]{@{}c@{}}$+0.9992$\\{[}$0.8679$, $1.1729${]}\end{tabular} \\
Lag-1 present, transition & \begin{tabular}[t]{@{}c@{}}$-0.0801$\\{[}$-0.0858$, $-0.0743${]}\end{tabular} & \begin{tabular}[t]{@{}c@{}}$-0.1328$\\{[}$-0.1659$, $-0.0993${]}\end{tabular} & \begin{tabular}[t]{@{}c@{}}$-0.2188$\\{[}$-0.2449$, $-0.1893${]}\end{tabular} \\
Lag-1 minus placebo, continuation & \begin{tabular}[t]{@{}c@{}}$+0.1910$\\{[}$0.1694$, $0.2127${]}\end{tabular} & \begin{tabular}[t]{@{}c@{}}$+0.7397$\\{[}$0.6097$, $0.8693${]}\end{tabular} & \begin{tabular}[t]{@{}c@{}}$+1.0527$\\{[}$0.9412$, $1.1755${]}\end{tabular} \\
Any same-dialogue exposure, continuation & \begin{tabular}[t]{@{}c@{}}$+0.0619$\\{[}$0.0557$, $0.0687${]}\end{tabular} & \begin{tabular}[t]{@{}c@{}}$+0.6407$\\{[}$0.5750$, $0.7104${]}\end{tabular} & \begin{tabular}[t]{@{}c@{}}$+1.2108$\\{[}$1.1450$, $1.2750${]}\end{tabular} \\
Any same-dialogue exposure, transition & \begin{tabular}[t]{@{}c@{}}$-0.1276$\\{[}$-0.1302$, $-0.1250${]}\end{tabular} & \begin{tabular}[t]{@{}c@{}}$-0.1820$\\{[}$-0.1942$, $-0.1693${]}\end{tabular} & \begin{tabular}[t]{@{}c@{}}$+0.1226$\\{[}$0.0788$, $0.1695${]}\end{tabular} \\
Row split minus grouped split, all targets & \begin{tabular}[t]{@{}c@{}}$-0.0903$\\{[}$-0.0928$, $-0.0878${]}\end{tabular} & \begin{tabular}[t]{@{}c@{}}$-0.0973$\\{[}$-0.1113$, $-0.0840${]}\end{tabular} & \begin{tabular}[t]{@{}c@{}}$+0.3842$\\{[}$0.3331$, $0.4370${]}\end{tabular} \\
Prespecified decision & present & present & present \\
\emph{TF-IDF classifier minus RoBERTa (exploratory)} &  &  &  \\
Lag-1 present, continuation, same targets (repeats 0--2) & \begin{tabular}[t]{@{}c@{}}$+0.0282$\\{[}$0.0113$, $0.0443${]}\end{tabular} & \begin{tabular}[t]{@{}c@{}}$+0.0844$\\{[}$0.0004$, $0.1705${]}\end{tabular} & \begin{tabular}[t]{@{}c@{}}$+0.2026$\\{[}$0.1432$, $0.2977${]}\end{tabular} \\
\bottomrule
\end{tabular}
\caption{Learner dependence. The encoder shows no detectable adjacency effect (inconclusive, not falsified; the SGD encoder is near ceiling); the $k$-NN learner has the largest effect on the bounded LRAP of SGD, but its log scores are unbounded below, so magnitudes are not comparable across learners on STAR and AMI. Last row: paired difference between the TF-IDF classifier and the encoder on the same targets (exploratory).}
\label{tab:a-learners}
\end{table*}

\section{Retrieval In-Context Learning: All Cells and Decision Rules}
\label{app:llm}

\paragraph{Conditions.} Exposed pool: other clusters plus the earlier labelled targets of the query's own dialogue.
Filtered pool: other clusters only. Label-matched replacement: every same-dialogue demonstration of the exposed list
replaced, in its slot, by the other-cluster window with the same label that is closest in similarity to the query and in
length; a second replacement uses a different label of nearest marginal frequency. Random: four random pool windows.
Zero-shot: no demonstration. On AMI, a weaker filter excludes only the query's meeting. A wrong-label condition, scored on
the fresh sample only, keeps the exposed list but shows each same-dialogue demonstration with the different label used by
the second replacement. The prespecified primary ordering (order A) places demonstrations in ascending similarity, with the most similar demonstration last. All primary and replication decisions were based on order A. A seeded per-query permutation of the four demonstration slots, applied consistently across conditions (order B), was evaluated as a sensitivity analysis rather than an alternative decision rule. The fresh SGD/Qwen transition loss was detectable under order B but not under order A; therefore, the prespecified replication decision remains partial.

\paragraph{Decision rules.} For the first sample, the effect was declared \emph{present} if the continuation gain and the
continuation-minus-transition interaction both had intervals above zero, and \emph{partial} if exactly one did. For
the fresh sample, the findings to recur were fixed per corpus from Qwen's first sample and applied unchanged to both
models: \emph{replicated} required all of them, and \emph{partial} meant that the interaction recurred but not all (for
SGD: the interaction, the transition loss, the label-placement contrast, and no positive same-dialogue increment beyond
label placement, but not the continuation gain, absent from Qwen's first sample; for AMI: the interaction, the
continuation gain and the same-dialogue increment). Table~\ref{tab:a-llm} lists all cells and
Table~\ref{tab:a-replication} the fresh-versus-first comparison. Free generation reproduced the AMI pattern, while 27--31\%
of SGD generations were not a valid act set, which is why constrained scoring is primary.

\begin{table*}[t]
\centering
\footnotesize
\setlength{\tabcolsep}{2.4pt}
\begin{tabular}{llrrrrrrrrrrrr}
\toprule
 &  & \multicolumn{3}{c}{Continuation (nats)} & \multicolumn{3}{c}{Transition (nats)} & \multicolumn{3}{c}{Interaction (nats)} & \multicolumn{3}{c}{Aggregate (top-1, natural)} \\
\cmidrule(lr){3-5} \cmidrule(lr){6-8} \cmidrule(lr){9-11} \cmidrule(lr){12-14}
Arm & Sample & est. & lo & hi & est. & lo & hi & est. & lo & hi & est. & lo & hi \\
\midrule
SGD Qwen & 1st & $+0.33$ & $-0.20$ & $0.87$ & $-0.86$ & $-1.40$ & $-0.36$ & $+1.19$ & $0.48$ & $1.92$ & $-0.027$ & $-0.057$ & $0.007$ \\
 & fresh & $+0.59$ & $0.06$ & $1.21$ & $-0.39$ & $-0.86$ & $0.06$ & $+0.98$ & $0.25$ & $1.74$ & $+0.004$ & $-0.026$ & $0.033$ \\
 & pooled & $+0.46$ & $0.08$ & $0.85$ & $-0.63$ & $-0.98$ & $-0.29$ & $+1.09$ & $0.60$ & $1.63$ & $-0.011$ & $-0.033$ & $0.011$ \\
\midrule
SGD Mistral & 1st & $+0.77$ & $0.30$ & $1.24$ & $-0.77$ & $-1.12$ & $-0.41$ & $+1.53$ & $0.92$ & $2.11$ & $-0.027$ & $-0.060$ & $0.007$ \\
 & fresh & $+0.33$ & $-0.06$ & $0.76$ & $-0.75$ & $-1.12$ & $-0.38$ & $+1.08$ & $0.55$ & $1.64$ & $-0.050$ & $-0.084$ & $-0.015$ \\
 & pooled & $+0.55$ & $0.25$ & $0.86$ & $-0.76$ & $-1.02$ & $-0.51$ & $+1.31$ & $0.93$ & $1.74$ & $-0.038$ & $-0.062$ & $-0.013$ \\
\midrule
AMI Qwen & 1st & $+2.43$ & $1.87$ & $3.00$ & $-0.31$ & $-0.92$ & $0.29$ & $+2.74$ & $1.99$ & $3.47$ & $+0.041$ & $0.018$ & $0.066$ \\
 & fresh & $+2.72$ & $2.10$ & $3.36$ & $-0.79$ & $-1.35$ & $-0.22$ & $+3.51$ & $2.58$ & $4.41$ & $+0.030$ & $0.006$ & $0.054$ \\
 & pooled & $+2.58$ & $2.17$ & $3.01$ & $-0.55$ & $-0.93$ & $-0.13$ & $+3.13$ & $2.54$ & $3.70$ & $+0.036$ & $0.017$ & $0.055$ \\
\midrule
AMI Mistral & 1st & $+2.16$ & $1.72$ & $2.64$ & $-0.74$ & $-1.16$ & $-0.29$ & $+2.89$ & $2.26$ & $3.54$ & $+0.027$ & $0.004$ & $0.051$ \\
 & fresh & $+2.63$ & $2.07$ & $3.21$ & $-0.98$ & $-1.50$ & $-0.47$ & $+3.62$ & $2.81$ & $4.36$ & $+0.013$ & $-0.008$ & $0.033$ \\
 & pooled & $+2.40$ & $1.99$ & $2.79$ & $-0.86$ & $-1.15$ & $-0.57$ & $+3.26$ & $2.76$ & $3.72$ & $+0.020$ & $0.004$ & $0.035$ \\
\midrule
STAR Qwen (expl.) & 1st & $+1.38$ & $0.91$ & $1.86$ & $+0.10$ & $-0.19$ & $0.41$ & $+1.28$ & $0.71$ & $1.81$ & $-0.019$ & $-0.049$ & $0.010$ \\
\bottomrule
\end{tabular}
\caption{All retrieval in-context learning cells: exposed minus filtered pool, demonstrations ordered by ascending similarity, $k=4$. 300 continuation and 450 transition queries per sample; pooled estimates use one joint cluster bootstrap over both samples.}
\label{tab:a-llm}
\end{table*}

\begin{table*}[t]
\centering
\footnotesize
\setlength{\tabcolsep}{4pt}
\begin{tabular}{lccccc}
\toprule
Arm & Decision & Compared & CI covers 0 & Same sign & Half-width, med./max \\
\midrule
SGD / Qwen & partial & 23 & 23 & 19 & 0.73 / 1.09 \\
SGD / Mistral & replicated & 19 & 19 & 19 & 0.56 / 0.84 \\
AMI / Qwen & replicated & 25 & 25 & 23 & 0.78 / 1.18 \\
AMI / Mistral & replicated & 19 & 19 & 18 & 0.55 / 1.04 \\
\bottomrule
\end{tabular}
\caption{Fresh-sample replication: fresh minus first-sample differences of all prespecified estimands. Intervals covering zero at these half-widths (nats) are not evidence of equivalence; no equivalence margin was fixed. AMI samples share all 36 meeting series.}
\label{tab:a-replication}
\end{table*}

\section{What the Leaked Window Carries}
\label{app:decomposition}

Table~\ref{tab:a-decomposition} decomposes the continuation gain. Placing the previous act's label on a label-matched
other-dialogue window raises continuation scores relative to a different label in all eight arms. The same-dialogue
increment beyond that replacement is negative on SGD and large on AMI, but Table~\ref{tab:a-support} shows that on AMI it
has no common support in similarity and grows with the similarity gap, so it is a natural retriever source-exposure
contrast rather than a provenance effect at fixed similarity. The wrong-label condition is not an unlabelled control: on
AMI it removes the copying (the transition copy rate falls from 0.380 under exposure to 0.171, against 0.140 with the
filter, for Qwen on the fresh sample), and its null aggregate effect is the sum of a positive text effect and a negative
wrong-label effect (Appendix~\ref{app:unlabelled}).

\begin{table*}[t]
\centering
\footnotesize
\setlength{\tabcolsep}{4pt}
\begin{tabular}{>{\raggedright\arraybackslash}p{4.3cm}lcccc}
\toprule
Contrast (continuations, nats) & Sample & SGD / Qwen & SGD / Mistral & AMI / Qwen & AMI / Mistral \\
\midrule
Previous act's label vs.\ different label, other-dialogue window & 1st & \begin{tabular}[t]{@{}c@{}}$+1.33$\\{[}$0.88$, $1.80${]}\end{tabular} & \begin{tabular}[t]{@{}c@{}}$+1.88$\\{[}$1.39$, $2.37${]}\end{tabular} & \begin{tabular}[t]{@{}c@{}}$+0.81$\\{[}$0.42$, $1.21${]}\end{tabular} & \begin{tabular}[t]{@{}c@{}}$+0.78$\\{[}$0.50$, $1.05${]}\end{tabular} \\
 & fresh & \begin{tabular}[t]{@{}c@{}}$+1.02$\\{[}$0.53$, $1.53${]}\end{tabular} & \begin{tabular}[t]{@{}c@{}}$+1.33$\\{[}$0.96$, $1.75${]}\end{tabular} & \begin{tabular}[t]{@{}c@{}}$+1.17$\\{[}$0.69$, $1.62${]}\end{tabular} & \begin{tabular}[t]{@{}c@{}}$+1.16$\\{[}$0.85$, $1.48${]}\end{tabular} \\
\midrule
Previous act's label on other-dialogue window vs.\ filtered & 1st & \begin{tabular}[t]{@{}c@{}}$+0.75$\\{[}$0.22$, $1.25${]}\end{tabular} & \begin{tabular}[t]{@{}c@{}}$+1.17$\\{[}$0.70$, $1.64${]}\end{tabular} & \begin{tabular}[t]{@{}c@{}}$-0.01$\\{[}$-0.32$, $0.25${]}\end{tabular} & \begin{tabular}[t]{@{}c@{}}$+0.71$\\{[}$0.42$, $1.00${]}\end{tabular} \\
 & fresh & \begin{tabular}[t]{@{}c@{}}$+0.86$\\{[}$0.34$, $1.45${]}\end{tabular} & \begin{tabular}[t]{@{}c@{}}$+1.07$\\{[}$0.71$, $1.46${]}\end{tabular} & \begin{tabular}[t]{@{}c@{}}$+0.12$\\{[}$-0.30$, $0.54${]}\end{tabular} & \begin{tabular}[t]{@{}c@{}}$+0.96$\\{[}$0.60$, $1.31${]}\end{tabular} \\
\midrule
Same-dialogue window vs.\ label-matched other window & 1st & \begin{tabular}[t]{@{}c@{}}$-0.41$\\{[}$-0.70$, $-0.14${]}\end{tabular} & \begin{tabular}[t]{@{}c@{}}$-0.40$\\{[}$-0.79$, $-0.03${]}\end{tabular} & \begin{tabular}[t]{@{}c@{}}$+2.45$\\{[}$1.89$, $3.05${]}\end{tabular} & \begin{tabular}[t]{@{}c@{}}$+1.45$\\{[}$1.00$, $1.92${]}\end{tabular} \\
 & fresh & \begin{tabular}[t]{@{}c@{}}$-0.27$\\{[}$-0.60$, $0.09${]}\end{tabular} & \begin{tabular}[t]{@{}c@{}}$-0.73$\\{[}$-1.02$, $-0.44${]}\end{tabular} & \begin{tabular}[t]{@{}c@{}}$+2.60$\\{[}$2.02$, $3.20${]}\end{tabular} & \begin{tabular}[t]{@{}c@{}}$+1.67$\\{[}$1.17$, $2.14${]}\end{tabular} \\
 & pooled & \begin{tabular}[t]{@{}c@{}}$-0.34$\\{[}$-0.57$, $-0.11${]}\end{tabular} & \begin{tabular}[t]{@{}c@{}}$-0.57$\\{[}$-0.82$, $-0.33${]}\end{tabular} & \begin{tabular}[t]{@{}c@{}}$+2.52$\\{[}$2.09$, $2.98${]}\end{tabular} & \begin{tabular}[t]{@{}c@{}}$+1.56$\\{[}$1.21$, $1.91${]}\end{tabular} \\
\midrule
Same-dialogue window, gold vs.\ wrong label & fresh & \begin{tabular}[t]{@{}c@{}}$+0.57$\\{[}$0.22$, $0.96${]}\end{tabular} & \begin{tabular}[t]{@{}c@{}}$0.00$\\{[}$-0.26$, $0.27${]}\end{tabular} & \begin{tabular}[t]{@{}c@{}}$+2.92$\\{[}$2.47$, $3.35${]}\end{tabular} & \begin{tabular}[t]{@{}c@{}}$+2.59$\\{[}$2.25$, $2.93${]}\end{tabular} \\
\midrule
Same-dialogue window with wrong label vs.\ filtered & fresh & \begin{tabular}[t]{@{}c@{}}$+0.02$\\{[}$-0.42$, $0.47${]}\end{tabular} & \begin{tabular}[t]{@{}c@{}}$+0.34$\\{[}$-0.09$, $0.77${]}\end{tabular} & \begin{tabular}[t]{@{}c@{}}$-0.20$\\{[}$-0.85$, $0.45${]}\end{tabular} & \begin{tabular}[t]{@{}c@{}}$+0.04$\\{[}$-0.47$, $0.57${]}\end{tabular} \\
\bottomrule
\end{tabular}
\caption{Decomposition of the continuation gain. Wrong-label contrasts exist only in the fresh sample. Label-matched replacements are on average 0.32 (SGD) and 0.44 (AMI) lower in cosine than the same-dialogue windows they replace, so the third contrast is not a provenance effect at fixed similarity (Table~\ref{tab:a-support}).}
\label{tab:a-decomposition}
\end{table*}

\begin{table*}[t]
\centering
\footnotesize
\setlength{\tabcolsep}{2pt}
\begin{tabular}{lcccccc}
\toprule
Arm & $[0,0.1]$ & $(0.1,0.2]$ & $(0.2,0.3]$ & $(0.3,0.5]$ & $>0.5$ & Slope \\
\midrule
SGD / Qwen & \begin{tabular}[t]{@{}c@{}}$-0.37$\\{[}$-1.04$, $0.29${]}\\$n=30$\end{tabular} & \begin{tabular}[t]{@{}c@{}}$-0.23$\\{[}$-0.56$, $0.06${]}\\$n=83$\end{tabular} & \begin{tabular}[t]{@{}c@{}}$-0.14$\\{[}$-0.82$, $0.49${]}\\$n=96$\end{tabular} & \begin{tabular}[t]{@{}c@{}}$-0.40$\\{[}$-0.77$, $-0.05${]}\\$n=265$\end{tabular} & \begin{tabular}[t]{@{}c@{}}$-0.49$\\{[}$-1.02$, $0.01${]}\\$n=110$\end{tabular} & \begin{tabular}[t]{@{}c@{}}$-0.59$\\{[}$-1.80$, $0.61${]}\end{tabular} \\
SGD / Mistral & \begin{tabular}[t]{@{}c@{}}$-1.53$\\{[}$-2.72$, $-0.48${]}\\$n=30$\end{tabular} & \begin{tabular}[t]{@{}c@{}}$-0.67$\\{[}$-1.35$, $-0.03${]}\\$n=83$\end{tabular} & \begin{tabular}[t]{@{}c@{}}$-0.48$\\{[}$-1.03$, $0.03${]}\\$n=96$\end{tabular} & \begin{tabular}[t]{@{}c@{}}$-0.33$\\{[}$-0.70$, $0.04${]}\\$n=265$\end{tabular} & \begin{tabular}[t]{@{}c@{}}$-0.96$\\{[}$-1.58$, $-0.38${]}\\$n=110$\end{tabular} & \begin{tabular}[t]{@{}c@{}}$+0.21$\\{[}$-1.38$, $1.78${]}\end{tabular} \\
AMI / Qwen & $n=0$ & \begin{tabular}[t]{@{}c@{}}$+0.05$\\{[}$-0.87$, $0.67${]}\\$n=13$\end{tabular} & \begin{tabular}[t]{@{}c@{}}$+1.67$\\{[}$-0.88$, $4.22${]}\\$n=18$\end{tabular} & \begin{tabular}[t]{@{}c@{}}$+2.76$\\{[}$1.89$, $3.62${]}\\$n=72$\end{tabular} & \begin{tabular}[t]{@{}c@{}}$+2.59$\\{[}$2.10$, $3.13${]}\\$n=495$\end{tabular} & \begin{tabular}[t]{@{}c@{}}$+3.25$\\{[}$0.61$, $5.92${]}\end{tabular} \\
AMI / Mistral & $n=0$ & \begin{tabular}[t]{@{}c@{}}$+0.99$\\{[}$0.27$, $1.63${]}\\$n=13$\end{tabular} & \begin{tabular}[t]{@{}c@{}}$+1.35$\\{[}$0.09$, $2.77${]}\\$n=18$\end{tabular} & \begin{tabular}[t]{@{}c@{}}$+1.77$\\{[}$0.82$, $2.67${]}\\$n=72$\end{tabular} & \begin{tabular}[t]{@{}c@{}}$+1.56$\\{[}$1.20$, $1.94${]}\\$n=495$\end{tabular} & \begin{tabular}[t]{@{}c@{}}$+1.95$\\{[}$0.25$, $3.67${]}\end{tabular} \\
\bottomrule
\end{tabular}
\caption{Common support of the same-dialogue increment (same-dialogue window minus label-matched other window, continuations, nats, both samples pooled) by the similarity gap between the two windows. The similarity gap is defined only for continuation queries with at least one same-dialogue demonstration in the exposed top-four retrieval. Of 600 continuation queries per corpus across both samples, 16 SGD and 2 AMI queries did not meet this condition, leaving 584 and 598 observations, respectively. AMI has no query with a gap of at most 0.1, and its increment grows with the gap.}
\label{tab:a-support}
\end{table*}

\section{Unlabelled Same-Dialogue Text}
\label{app:unlabelled}

The control of Table~\ref{tab:a-unlabelled} was designed after the wrong-label result and is exploratory; its validity
checks (identical labelled demonstrations, prompt lengths within 25 tokens, and a positive effect of the labelled
same-dialogue window) were met in all four arms, and an independent reimplementation reproduced all of its estimates.
The follow-up of Table~\ref{tab:a-window} used Qwen only and new queries that the control had not used (100
continuations and 100 transitions per corpus), split the lag-1 window by corpus position into the utterances inside the
query window and the older ones, and compared each with a token-matched excerpt from another dialogue. On SGD, 47\% of
continuation queries have no older utterance because their window starts at the dialogue opening, so the older-part
contrast rests on 53 continuation queries. On AMI, 52.5\% of queries render at least one shared utterance with a
different speaker tag than in the query window, as AMI speaker tags are relative to the target's speaker.

\begin{table*}[t]
\centering
\footnotesize
\setlength{\tabcolsep}{2.5pt}
\begin{tabular}{>{\raggedright\arraybackslash}p{4.1cm}cccc}
\toprule
Contrast (continuations, nats) & SGD / Qwen & SGD / Mistral & AMI / Qwen & AMI / Mistral \\
\midrule
Unlabelled same-dialogue vs.\ unlabelled other window & \begin{tabular}[t]{@{}c@{}}$+0.43$\\{[}$0.12$, $0.77${]}\end{tabular} & \begin{tabular}[t]{@{}c@{}}$+1.07$\\{[}$0.64$, $1.53${]}\end{tabular} & \begin{tabular}[t]{@{}c@{}}$+1.14$\\{[}$0.64$, $1.67${]}\end{tabular} & \begin{tabular}[t]{@{}c@{}}$+0.75$\\{[}$0.32$, $1.16${]}\end{tabular} \\
Unlabelled same-dialogue window vs.\ none & \begin{tabular}[t]{@{}c@{}}$+0.50$\\{[}$0.16$, $0.86${]}\end{tabular} & \begin{tabular}[t]{@{}c@{}}$+1.42$\\{[}$0.90$, $1.95${]}\end{tabular} & \begin{tabular}[t]{@{}c@{}}$+1.50$\\{[}$1.00$, $2.05${]}\end{tabular} & \begin{tabular}[t]{@{}c@{}}$+0.88$\\{[}$0.42$, $1.29${]}\end{tabular} \\
Unlabelled other window vs.\ none & \begin{tabular}[t]{@{}c@{}}$+0.07$\\{[}$-0.31$, $0.43${]}\end{tabular} & \begin{tabular}[t]{@{}c@{}}$+0.35$\\{[}$-0.01$, $0.70${]}\end{tabular} & \begin{tabular}[t]{@{}c@{}}$+0.37$\\{[}$-0.07$, $0.80${]}\end{tabular} & \begin{tabular}[t]{@{}c@{}}$+0.13$\\{[}$-0.17$, $0.41${]}\end{tabular} \\
Gold label vs.\ no label, same text & \begin{tabular}[t]{@{}c@{}}$+0.31$\\{[}$-0.10$, $0.74${]}\end{tabular} & \begin{tabular}[t]{@{}c@{}}$-0.54$\\{[}$-1.05$, $-0.02${]}\end{tabular} & \begin{tabular}[t]{@{}c@{}}$+2.00$\\{[}$1.55$, $2.42${]}\end{tabular} & \begin{tabular}[t]{@{}c@{}}$+1.61$\\{[}$1.21$, $2.06${]}\end{tabular} \\
Wrong label vs.\ no label, same text & \begin{tabular}[t]{@{}c@{}}$-0.08$\\{[}$-0.36$, $0.19${]}\end{tabular} & \begin{tabular}[t]{@{}c@{}}$-1.03$\\{[}$-1.48$, $-0.62${]}\end{tabular} & \begin{tabular}[t]{@{}c@{}}$-1.65$\\{[}$-2.15$, $-1.16${]}\end{tabular} & \begin{tabular}[t]{@{}c@{}}$-0.51$\\{[}$-0.77$, $-0.27${]}\end{tabular} \\
Labelled same-dialogue window vs.\ none & \begin{tabular}[t]{@{}c@{}}$+0.82$\\{[}$0.40$, $1.25${]}\end{tabular} & \begin{tabular}[t]{@{}c@{}}$+0.87$\\{[}$0.36$, $1.40${]}\end{tabular} & \begin{tabular}[t]{@{}c@{}}$+3.51$\\{[}$2.90$, $4.09${]}\end{tabular} & \begin{tabular}[t]{@{}c@{}}$+2.49$\\{[}$1.87$, $3.13${]}\end{tabular} \\
\midrule
Previous-act mass, unlabelled same-dialogue, continuation & \begin{tabular}[t]{@{}c@{}}$+0.050$\\{[}$0.008$, $0.092${]}\end{tabular} & \begin{tabular}[t]{@{}c@{}}$+0.104$\\{[}$0.064$, $0.147${]}\end{tabular} & \begin{tabular}[t]{@{}c@{}}$+0.096$\\{[}$0.053$, $0.144${]}\end{tabular} & \begin{tabular}[t]{@{}c@{}}$+0.065$\\{[}$0.017$, $0.112${]}\end{tabular} \\
Previous-act mass, unlabelled same-dialogue, transition & \begin{tabular}[t]{@{}c@{}}$-0.006$\\{[}$-0.042$, $0.028${]}\end{tabular} & \begin{tabular}[t]{@{}c@{}}$+0.060$\\{[}$0.029$, $0.097${]}\end{tabular} & \begin{tabular}[t]{@{}c@{}}$+0.054$\\{[}$0.020$, $0.085${]}\end{tabular} & \begin{tabular}[t]{@{}c@{}}$+0.029$\\{[}$0.005$, $0.054${]}\end{tabular} \\
Previous-act mass, unlabelled other, continuation & \begin{tabular}[t]{@{}c@{}}$+0.015$\\{[}$-0.032$, $0.060${]}\end{tabular} & \begin{tabular}[t]{@{}c@{}}$+0.031$\\{[}$-0.002$, $0.066${]}\end{tabular} & \begin{tabular}[t]{@{}c@{}}$+0.023$\\{[}$-0.007$, $0.054${]}\end{tabular} & \begin{tabular}[t]{@{}c@{}}$+0.006$\\{[}$-0.026$, $0.041${]}\end{tabular} \\
Previous-act mass, unlabelled other, transition & \begin{tabular}[t]{@{}c@{}}$-0.010$\\{[}$-0.041$, $0.021${]}\end{tabular} & \begin{tabular}[t]{@{}c@{}}$-0.005$\\{[}$-0.032$, $0.025${]}\end{tabular} & \begin{tabular}[t]{@{}c@{}}$-0.008$\\{[}$-0.033$, $0.017${]}\end{tabular} & \begin{tabular}[t]{@{}c@{}}$-0.011$\\{[}$-0.035$, $0.013${]}\end{tabular} \\
\bottomrule
\end{tabular}
\caption{Exploratory unlabelled-context control (150 continuation and 150 transition queries per corpus). Every arm holds the same three labelled other-dialogue demonstrations. The matched other window is 0.36 (SGD) and 0.62 (AMI) lower in cosine than the lag-1 window, so the first row is not a contrast at equal similarity. Lower block: posterior mass on the previous act, arm minus no extra block.}
\label{tab:a-unlabelled}
\end{table*}

\begin{table*}[t]
\centering
\footnotesize
\setlength{\tabcolsep}{2.5pt}
\begin{tabular}{>{\raggedright\arraybackslash}p{3.2cm}cccc}
\toprule
Contrast (nats) & SGD cont. & SGD trans. & AMI cont. & AMI trans. \\
\midrule
Shared part vs.\ matched other text & \begin{tabular}[t]{@{}c@{}}$+0.23$\\{[}$-0.15$, $0.59${]}\\$n=100$\end{tabular} & \begin{tabular}[t]{@{}c@{}}$-0.52$\\{[}$-1.18$, $0.03${]}\\$n=100$\end{tabular} & \begin{tabular}[t]{@{}c@{}}$+0.83$\\{[}$0.16$, $1.54${]}\\$n=100$\end{tabular} & \begin{tabular}[t]{@{}c@{}}$+0.46$\\{[}$-0.09$, $1.02${]}\\$n=100$\end{tabular} \\
Older part vs.\ matched other text & \begin{tabular}[t]{@{}c@{}}$-0.29$\\{[}$-0.74$, $0.17${]}\\$n=53$\end{tabular} & \begin{tabular}[t]{@{}c@{}}$-0.46$\\{[}$-0.92$, $-0.04${]}\\$n=82$\end{tabular} & \begin{tabular}[t]{@{}c@{}}$+0.11$\\{[}$-0.18$, $0.39${]}\\$n=100$\end{tabular} & \begin{tabular}[t]{@{}c@{}}$-0.17$\\{[}$-0.63$, $0.29${]}\\$n=100$\end{tabular} \\
Whole window vs.\ empty block & \begin{tabular}[t]{@{}c@{}}$+0.13$\\{[}$-0.24$, $0.53${]}\\$n=100$\end{tabular} & \begin{tabular}[t]{@{}c@{}}$-0.06$\\{[}$-0.72$, $0.54${]}\\$n=100$\end{tabular} & \begin{tabular}[t]{@{}c@{}}$+0.69$\\{[}$-0.13$, $1.43${]}\\$n=100$\end{tabular} & \begin{tabular}[t]{@{}c@{}}$+0.41$\\{[}$-0.29$, $1.07${]}\\$n=100$\end{tabular} \\
Shared part vs.\ empty block & \begin{tabular}[t]{@{}c@{}}$+0.23$\\{[}$-0.11$, $0.57${]}\\$n=100$\end{tabular} & \begin{tabular}[t]{@{}c@{}}$-0.16$\\{[}$-0.79$, $0.43${]}\\$n=100$\end{tabular} & \begin{tabular}[t]{@{}c@{}}$+0.89$\\{[}$0.19$, $1.58${]}\\$n=100$\end{tabular} & \begin{tabular}[t]{@{}c@{}}$+0.27$\\{[}$-0.37$, $0.84${]}\\$n=100$\end{tabular} \\
Older part vs.\ empty block & \begin{tabular}[t]{@{}c@{}}$+0.10$\\{[}$-0.43$, $0.69${]}\\$n=53$\end{tabular} & \begin{tabular}[t]{@{}c@{}}$-0.21$\\{[}$-0.73$, $0.29${]}\\$n=82$\end{tabular} & \begin{tabular}[t]{@{}c@{}}$+0.28$\\{[}$-0.08$, $0.64${]}\\$n=100$\end{tabular} & \begin{tabular}[t]{@{}c@{}}$+0.25$\\{[}$-0.21$, $0.68${]}\\$n=100$\end{tabular} \\
Whole window vs.\ shared part & \begin{tabular}[t]{@{}c@{}}$-0.19$\\{[}$-0.63$, $0.23${]}\\$n=53$\end{tabular} & \begin{tabular}[t]{@{}c@{}}$+0.12$\\{[}$-0.25$, $0.51${]}\\$n=82$\end{tabular} & \begin{tabular}[t]{@{}c@{}}$-0.21$\\{[}$-0.63$, $0.21${]}\\$n=100$\end{tabular} & \begin{tabular}[t]{@{}c@{}}$+0.14$\\{[}$-0.29$, $0.60${]}\\$n=100$\end{tabular} \\
Matched other text vs.\ empty block & \begin{tabular}[t]{@{}c@{}}$0.00$\\{[}$-0.33$, $0.36${]}\\$n=100$\end{tabular} & \begin{tabular}[t]{@{}c@{}}$+0.36$\\{[}$-0.08$, $0.86${]}\\$n=100$\end{tabular} & \begin{tabular}[t]{@{}c@{}}$+0.07$\\{[}$-0.34$, $0.46${]}\\$n=100$\end{tabular} & \begin{tabular}[t]{@{}c@{}}$-0.19$\\{[}$-0.63$, $0.22${]}\\$n=100$\end{tabular} \\
\midrule
AMI shared vs.\ other, speaker tags differ & -- & -- & \begin{tabular}[t]{@{}c@{}}$+1.80$\\{[}$0.63$, $3.13${]}\\$n=43$\end{tabular} & -- \\
AMI shared vs.\ other, speaker tags agree & -- & -- & \begin{tabular}[t]{@{}c@{}}$+0.10$\\{[}$-0.61$, $0.86${]}\\$n=57$\end{tabular} & -- \\
Previous-act mass, shared part $-$ empty & \begin{tabular}[t]{@{}c@{}}$+0.022$\\{[}$-0.015$, $0.065${]}\end{tabular} & \begin{tabular}[t]{@{}c@{}}$+0.036$\\{[}$-0.006$, $0.083${]}\end{tabular} & \begin{tabular}[t]{@{}c@{}}$+0.021$\\{[}$-0.043$, $0.088${]}\end{tabular} & \begin{tabular}[t]{@{}c@{}}$+0.044$\\{[}$-0.015$, $0.105${]}\end{tabular} \\
\bottomrule
\end{tabular}
\caption{Follow-up decomposition of the unlabelled lag-1 window into the part already inside the query window and the older part (one model; new queries not used in the control above). Placebos are token-matched excerpts from other dialogues; all length checks passed. The speaker-tag strata are descriptive and not randomised.}
\label{tab:a-window}
\end{table*}

\section{Filtering and Previous-Act Predictions}
\label{app:filter}

\begin{table*}[t]
\centering
\begingroup
\includegraphics[width=\textwidth]{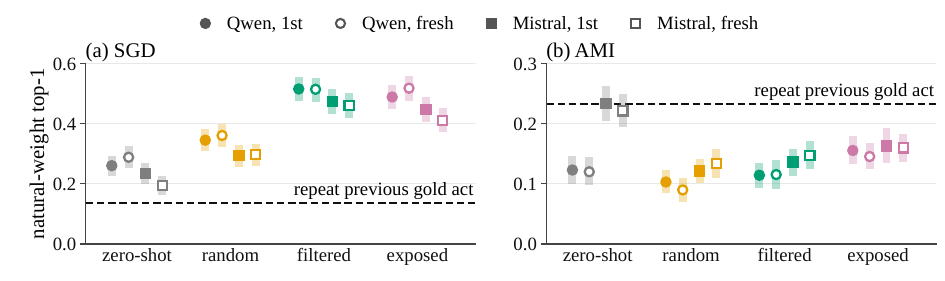}
\captionof{figure}{Cost of the filter: natural-weight top-1 accuracy by demonstration condition (categories, not a sequence) for
each model and sample, with 95\% cluster-bootstrap intervals (bands); dashed: repeating the previous gold act, a baseline
in Setting B, inadmissible in Setting A. Values: Table~\ref{tab:a-filter}.}
\label{fig:utility}
\endgroup
\medskip

\footnotesize
\setlength{\tabcolsep}{3pt}
\begin{tabular}{llrrrrrr}
\toprule
 &  & \multicolumn{3}{c}{Qwen} & \multicolumn{3}{c}{Mistral} \\
\cmidrule(lr){3-5} \cmidrule(lr){6-8}
Condition & Sample & est. & lo & hi & est. & lo & hi \\
\midrule
\multicolumn{8}{l}{\emph{SGD}} \\
Zero-shot & 1st & $0.261$ & $0.226$ & $0.294$ & $0.234$ & $0.199$ & $0.269$ \\
 & fresh & $0.289$ & $0.253$ & $0.326$ & $0.195$ & $0.164$ & $0.228$ \\
Random demonstrations & 1st & $0.346$ & $0.308$ & $0.384$ & $0.294$ & $0.258$ & $0.330$ \\
 & fresh & $0.361$ & $0.322$ & $0.401$ & $0.298$ & $0.260$ & $0.334$ \\
Filtered retrieval & 1st & $0.516$ & $0.476$ & $0.556$ & $0.474$ & $0.432$ & $0.516$ \\
 & fresh & $0.515$ & $0.473$ & $0.554$ & $0.461$ & $0.421$ & $0.502$ \\
Exposed retrieval & 1st & $0.489$ & $0.449$ & $0.531$ & $0.447$ & $0.407$ & $0.488$ \\
 & fresh & $0.519$ & $0.477$ & $0.559$ & $0.412$ & $0.371$ & $0.452$ \\
Repeat the previous act & -- & \multicolumn{3}{c}{0.135} & \multicolumn{3}{c}{0.135} \\
Filtered $-$ zero-shot & 1st & $+0.255$ & $0.217$ & $0.294$ & $+0.239$ & $0.192$ & $0.288$ \\
 & fresh & $+0.226$ & $0.187$ & $0.266$ & $+0.266$ & $0.221$ & $0.313$ \\
Filtered $-$ random & 1st & $+0.171$ & $0.128$ & $0.210$ & $+0.179$ & $0.133$ & $0.225$ \\
 & fresh & $+0.154$ & $0.113$ & $0.195$ & $+0.163$ & $0.123$ & $0.206$ \\
Exposed $-$ filtered & 1st & $-0.027$ & $-0.057$ & $0.007$ & $-0.027$ & $-0.060$ & $0.007$ \\
 & fresh & $+0.004$ & $-0.026$ & $0.033$ & $-0.050$ & $-0.084$ & $-0.015$ \\
\midrule
\multicolumn{8}{l}{\emph{AMI}} \\
Zero-shot & 1st & $0.123$ & $0.100$ & $0.146$ & $0.233$ & $0.205$ & $0.263$ \\
 & fresh & $0.120$ & $0.098$ & $0.144$ & $0.222$ & $0.194$ & $0.250$ \\
Random demonstrations & 1st & $0.103$ & $0.085$ & $0.123$ & $0.122$ & $0.102$ & $0.141$ \\
 & fresh & $0.090$ & $0.070$ & $0.110$ & $0.134$ & $0.109$ & $0.157$ \\
Filtered retrieval & 1st & $0.114$ & $0.094$ & $0.135$ & $0.136$ & $0.113$ & $0.159$ \\
 & fresh & $0.116$ & $0.091$ & $0.140$ & $0.148$ & $0.124$ & $0.172$ \\
Exposed retrieval & 1st & $0.156$ & $0.133$ & $0.179$ & $0.163$ & $0.135$ & $0.192$ \\
 & fresh & $0.145$ & $0.125$ & $0.168$ & $0.160$ & $0.136$ & $0.183$ \\
Repeat the previous act & -- & \multicolumn{3}{c}{0.232} & \multicolumn{3}{c}{0.232} \\
Filtered $-$ zero-shot & 1st & $-0.009$ & $-0.039$ & $0.022$ & $-0.097$ & $-0.131$ & $-0.065$ \\
 & fresh & $-0.005$ & $-0.033$ & $0.024$ & $-0.074$ & $-0.101$ & $-0.046$ \\
Filtered $-$ random & 1st & $+0.011$ & $-0.008$ & $0.032$ & $+0.014$ & $-0.009$ & $0.037$ \\
 & fresh & $+0.026$ & $-0.001$ & $0.052$ & $+0.014$ & $-0.007$ & $0.037$ \\
Exposed $-$ filtered & 1st & $+0.041$ & $0.018$ & $0.066$ & $+0.027$ & $0.004$ & $0.051$ \\
 & fresh & $+0.030$ & $0.006$ & $0.054$ & $+0.013$ & $-0.008$ & $0.033$ \\
\bottomrule
\end{tabular}
\caption{Utility of the provenance filter: top-1 accuracy at natural cell weights (levels, then paired differences) with 95\% cluster-bootstrap bounds. Repeating the previous act scores 1 on continuations and 0 on transitions, so its natural-weight accuracy equals the continuation share. It is an admissible model only when the previous act is genuinely observed.}
\label{tab:a-filter}

\end{table*}

Table~\ref{tab:a-filter} gives the accuracy levels behind Figure~\ref{fig:utility}, and Table~\ref{tab:a-copy} the
previous-act copy rates on transitions for every condition. For a condition with demonstrations $D$, the copy rate is
$\mathrm{CR}^{(D)}=|Q_T|^{-1}\sum_{q\in Q_T}\mathbf{1}[\hat y_q^{(D)}=y_{p(q)}]$ over the transition queries $Q_T$
($|Q_T|=450$ per sample), where $\hat y_q^{(D)}$ is the top-1 prediction; on a transition such a prediction is always
wrong. Table~\ref{tab:a-copy} gives the paired difference exposed minus filtered with its 95\% cluster-bootstrap
interval (Figure~\ref{fig:copy}).

\begin{table*}[t]
\centering
\footnotesize
\setlength{\tabcolsep}{3pt}
\begin{tabular}{llccccccrrr}
\toprule
 & & \multicolumn{6}{c}{Copy rate} & \multicolumn{3}{c}{Exposed $-$ filtered} \\ \cmidrule(lr){3-8} \cmidrule(lr){9-11}
Arm & Sample & Zero-shot & Random & Filtered & Exposed & Label-matched & Wrong label & est. & lo & hi \\
\midrule
SGD Qwen & 1st & 0.264 & 0.269 & 0.100 & 0.113 & 0.127 & -- & $+0.013$ & $-0.018$ & $0.044$ \\
 & fresh & 0.251 & 0.287 & 0.129 & 0.129 & 0.149 & 0.140 & $0.000$ & $-0.033$ & $0.032$ \\
SGD Mistral & 1st & 0.180 & 0.198 & 0.102 & 0.127 & 0.133 & -- & $+0.024$ & $-0.008$ & $0.057$ \\
 & fresh & 0.164 & 0.227 & 0.113 & 0.158 & 0.140 & 0.182 & $+0.044$ & $0.011$ & $0.082$ \\
\midrule
AMI Qwen & 1st & 0.151 & 0.091 & 0.102 & 0.336 & 0.129 & -- & $+0.233$ & $0.184$ & $0.283$ \\
 & fresh & 0.176 & 0.136 & 0.140 & 0.380 & 0.204 & 0.171 & $+0.240$ & $0.188$ & $0.294$ \\
AMI Mistral & 1st & 0.218 & 0.122 & 0.124 & 0.353 & 0.189 & -- & $+0.229$ & $0.190$ & $0.269$ \\
 & fresh & 0.240 & 0.153 & 0.147 & 0.371 & 0.198 & 0.160 & $+0.224$ & $0.191$ & $0.261$ \\
\bottomrule
\end{tabular}
\caption{Share of the 450 transition queries per sample predicted as the previous act (always wrong on transitions), and the paired difference exposed minus filtered with 95\% cluster-bootstrap bounds. The wrong-label condition exists only in the fresh sample.}
\label{tab:a-copy}
\end{table*}

\section{Retrieval Prevalence under Two Pipelines}
\label{app:prevalence}

Under TF-IDF retrieval, the lag-1 window is far more similar to the query than any other candidate: on fresh
continuation queries its mean cosine is 0.63 (SGD) and 0.79 (AMI), against 0.36 and 0.22 for the most similar
other-cluster window of any label. On average 72\% (SGD) and 80\% (AMI) of the lag-1 window's utterances are already
inside the query window, and the utterance whose act is the lag-1 label is inside the query window for every target. The
dense pipeline embeds the same rendered history text that the language model sees, with [CLS] pooling, L2 normalisation
and no instruction; it truncated no window. The TF-IDF pipeline uses the classifier's structural token sequence, with
role and window-length tokens; Table~\ref{tab:a-prevalence} gives all shares.

\begin{table*}[t]
\centering
\footnotesize
\setlength{\tabcolsep}{3pt}
\begin{tabular}{llrrrrrrrrr}
\toprule
 &  & \multicolumn{3}{c}{TF-IDF pipeline} & \multicolumn{3}{c}{Dense pipeline} & \multicolumn{3}{c}{Dense $-$ TF-IDF} \\
\cmidrule(lr){3-5} \cmidrule(lr){6-8} \cmidrule(lr){9-11}
Sample & Share of queries & est. & lo & hi & est. & lo & hi & est. & lo & hi \\
\midrule
SGD, 1st & Lag-1 in top 4 & $0.976$ & $0.964$ & $0.987$ & $0.432$ & $0.397$ & $0.467$ & $-0.544$ & $-0.580$ & $-0.509$ \\
 & Lag-1 nearest & $0.931$ & $0.911$ & $0.948$ & $0.309$ & $0.277$ & $0.343$ & $-0.621$ & $-0.656$ & $-0.585$ \\
 & Any same-dialogue & $0.976$ & $0.964$ & $0.987$ & $0.436$ & $0.400$ & $0.472$ & $-0.540$ & $-0.576$ & $-0.504$ \\
 & Top 4, continuations & $0.983$ & $0.968$ & $0.997$ & $0.407$ & $0.351$ & $0.463$ & $-0.577$ & $-0.632$ & $-0.520$ \\
 & Top 4, transitions & $0.971$ & $0.955$ & $0.986$ & $0.449$ & $0.403$ & $0.493$ & $-0.522$ & $-0.568$ & $-0.477$ \\
\midrule
SGD, fresh & Lag-1 in top 4 & $0.975$ & $0.963$ & $0.984$ & $0.429$ & $0.396$ & $0.464$ & $-0.545$ & $-0.580$ & $-0.508$ \\
 & Lag-1 nearest & $0.931$ & $0.912$ & $0.948$ & $0.321$ & $0.289$ & $0.356$ & $-0.609$ & $-0.644$ & $-0.573$ \\
 & Any same-dialogue & $0.975$ & $0.963$ & $0.984$ & $0.437$ & $0.404$ & $0.473$ & $-0.537$ & $-0.572$ & $-0.500$ \\
 & Top 4, continuations & $0.963$ & $0.939$ & $0.982$ & $0.390$ & $0.336$ & $0.449$ & $-0.573$ & $-0.629$ & $-0.516$ \\
 & Top 4, transitions & $0.982$ & $0.970$ & $0.993$ & $0.456$ & $0.410$ & $0.500$ & $-0.527$ & $-0.572$ & $-0.481$ \\
\midrule
AMI, 1st & Lag-1 in top 4 & $0.993$ & $0.986$ & $0.999$ & $0.839$ & $0.813$ & $0.862$ & $-0.155$ & $-0.182$ & $-0.130$ \\
 & Lag-1 nearest & $0.955$ & $0.937$ & $0.970$ & $0.703$ & $0.672$ & $0.730$ & $-0.252$ & $-0.284$ & $-0.222$ \\
 & Any same-dialogue & $0.995$ & $0.988$ & $1.000$ & $0.871$ & $0.845$ & $0.894$ & $-0.124$ & $-0.149$ & $-0.102$ \\
 & Top 4, continuations & $0.997$ & $0.990$ & $1.000$ & $0.830$ & $0.799$ & $0.862$ & $-0.167$ & $-0.199$ & $-0.134$ \\
 & Top 4, transitions & $0.991$ & $0.982$ & $0.998$ & $0.844$ & $0.807$ & $0.877$ & $-0.147$ & $-0.184$ & $-0.113$ \\
\midrule
AMI, fresh & Lag-1 in top 4 & $0.992$ & $0.986$ & $0.997$ & $0.839$ & $0.811$ & $0.866$ & $-0.153$ & $-0.180$ & $-0.128$ \\
 & Lag-1 nearest & $0.948$ & $0.933$ & $0.963$ & $0.696$ & $0.658$ & $0.732$ & $-0.252$ & $-0.286$ & $-0.222$ \\
 & Any same-dialogue & $0.995$ & $0.989$ & $0.999$ & $0.864$ & $0.838$ & $0.890$ & $-0.131$ & $-0.156$ & $-0.105$ \\
 & Top 4, continuations & $0.997$ & $0.990$ & $1.000$ & $0.840$ & $0.790$ & $0.891$ & $-0.157$ & $-0.206$ & $-0.108$ \\
 & Top 4, transitions & $0.989$ & $0.980$ & $0.998$ & $0.838$ & $0.805$ & $0.869$ & $-0.151$ & $-0.182$ & $-0.124$ \\
\midrule
STAR, 1st & Lag-1 in top 4 & $0.756$ & $0.727$ & $0.786$ & $0.272$ & $0.241$ & $0.302$ & $-0.484$ & $-0.521$ & $-0.450$ \\
 & Lag-1 nearest & $0.672$ & $0.641$ & $0.705$ & $0.183$ & $0.156$ & $0.210$ & $-0.489$ & $-0.523$ & $-0.454$ \\
 & Any same-dialogue & $0.760$ & $0.731$ & $0.790$ & $0.279$ & $0.247$ & $0.310$ & $-0.481$ & $-0.517$ & $-0.446$ \\
 & Top 4, continuations & $0.770$ & $0.720$ & $0.821$ & $0.260$ & $0.210$ & $0.309$ & $-0.510$ & $-0.569$ & $-0.454$ \\
 & Top 4, transitions & $0.747$ & $0.709$ & $0.786$ & $0.280$ & $0.239$ & $0.321$ & $-0.467$ & $-0.512$ & $-0.423$ \\
\bottomrule
\end{tabular}
\caption{Retrieval structure under two pipelines (750 queries per sample; SGD and STAR clusters are dialogues or near-duplicate groups, AMI clusters are meeting series), with 95\% cluster-bootstrap bounds. For SGD and AMI, the dense-only lag-1 retrieval count was zero in both the first and fresh samples (0 of 750 queries per corpus and sample). No language model was run on dense-retrieved demonstrations.}
\label{tab:a-prevalence}
\end{table*}

\section{Negative, Partial and Withdrawn Results}
\label{app:negative}

\begin{table*}[t]
\centering
\footnotesize
\setlength{\tabcolsep}{3pt}
\begin{tabular}{>{\raggedright\arraybackslash}p{4.6cm}>{\raggedright\arraybackslash}p{3.6cm}>{\raggedright\arraybackslash}p{6.8cm}}
\toprule
Prediction, control or reading & Outcome & Evidence \\
\midrule
Lead-1 presence helps transitions & not observed in 3 of 3 corpora & SGD $-0.0039$ [$-0.0074$, $-0.0006$]; STAR $-0.0376$ [$-0.0426$, $-0.0324$]; AMI $-0.0051$ [$-0.0209$, $0.0056$] \\
Masking the target text is a neutral control & not neutral in 3 of 3 & masked minus placebo-masked gain; Appendix~\ref{app:randomised} \\
Fine-tuned encoder shows an adjacency effect & inconclusive in 3 of 3 & Table~\ref{tab:a-learners} \\
SGD continuation gain, Qwen, first sample & missed; decision partial & $+0.33$ [$-0.20$, $0.87$] \\
SGD fresh replication, Qwen & partial: transition loss missed & $-0.39$ [$-0.86$, $0.06$] \\
Label agreement of the retrieved neighbour predicts the transition sign & unstable (1 of 4 fresh arms) & no claim \\
Same-dialogue text without its label adds nothing & withdrawn: that condition showed a wrong label & Table~\ref{tab:a-unlabelled} \\
AMI same-meeting increment as provenance at fixed similarity & not identified (no common support) & Table~\ref{tab:a-support} \\
Shared and older parts of the SGD window & no detectable effect & $+0.23$ [$-0.15$, $0.59$]; $-0.29$ [$-0.74$, $0.17$] \\
Unlabelled text acts as a persistence cue & not confirmed in the follow-up & Table~\ref{tab:a-window} \\
Independent assessment of a published row-split system & not feasible & Appendix~\ref{app:external} \\
\bottomrule
\end{tabular}
\caption{Negative, partial and withdrawn results. Each remains in the analyses; none was dropped.}
\label{tab:a-negative}
\end{table*}

Table~\ref{tab:a-negative} lists every failed prediction, every control that did not behave as intended and every
withdrawn reading; all remain in the analyses above.

\section{Feasibility of Evaluating a Published System}
\label{app:external}

The closest published next-act study with a confirmed window-level split is \citet{tanaka2019dialogue}: 1,155 SwDA
conversations (219,297 utterances) cut into 212,367 sliding five-utterance windows, split at random 80/10/10 at window
level (the paper calls the windows ``conversations''; their count and the 1,061,835 utterances they contain identify
them as windows), with the gold acts of earlier turns given to the model at test time by design. Estimating a split-dependent effect
would require a lawful copy of the corpus, the original implementation, its split and tag mapping, and a matched
row-versus-dialogue-disjoint evaluation; it was not feasible. No code or supplementary material accompanies the paper,
no implementation by its authors could be found, the split seed and indices and the mapping from the original tags to
its nine groups are not published, and no copy of the corpus was available locally. We therefore do not estimate how much, if at all, its reported
accuracy (69.7\%) or macro-F1 (32.4\%) depends on the split.

What the published protocol class implies can be computed without data. If windows slide by one utterance, are assigned
to training uniformly at random with probability 0.8, and the gold acts of a window's five input utterances are model
inputs, then an interior test window has its lag-1 labelled window in training with probability 0.8, and the gold act of
its own target appears among the inputs of at least one training window (the five following windows contain the target
utterance) with probability $1-0.2^5=0.99968$. These are exposure rates, not a measured effect.

\section{Reproducibility, Data and Compute}
\label{app:repro}

Confirmatory analyses were specified before model execution, with data selections, query sets, demonstration plans, and code fingerprints recorded. Exploratory analyses are identified separately. Experiments used pinned model revisions and documented configurations. The public repository provides the experimental code, reproduction instructions, independent checking scripts, and a mapping between reported results and their corresponding analyses. The experiments used a single 32\,GB GPU, with detailed computational requirements and model revisions documented in the repository.

Code and reproduction instructions are publicly available at
\url{https://github.com/AliAmini93/next-act-exposure-repro}.
Datasets and pretrained models must be obtained separately under their respective licences; corpus-derived records and model weights are not redistributed.

\end{document}